\documentclass[journal]{IEEEtran}
\IEEEoverridecommandlockouts
\usepackage[utf8]{inputenc}
\usepackage{multicol}
\usepackage{amssymb,graphicx,epstopdf,amsmath,color,algorithm,tabularx,float,url,gensymb,bm}
\usepackage[noadjust]{cite} % groups citations
\usepackage{subcaption,overpic}
\usepackage{placeins}
\usepackage{hyperref,dblfloatfix,verbatim}
\usepackage{cleveref}
\usepackage{multirow}
\usepackage{hhline}
\usepackage{varwidth}
\usepackage{tikz}
\usepackage{pict2e}
\usepackage{wrapfig}
\usepackage{xcolor, soul}
\usepackage[T1]{fontenc} % Fixes the issue of £ showing up as $

\newcommand\copyrighttext{%
  \footnotesize \copyright 2020 IEEE. Permission from IEEE must be obtained for all other uses, in any current or future media, including reprinting/republishing this material for advertising or promotional purposes, creating new collective works, for resale or redistribution to servers or lists, or reuse of any copyrighted component of this work in other works.}

\newcommand\copyrightnotice{%
\begin{tikzpicture}[remember picture,overlay]
\node[anchor=south,yshift=10pt] at (current page.south) {\fbox{\parbox{\dimexpr\textwidth-\fboxsep-\fboxrule\relax}{\copyrighttext}}};
\end{tikzpicture}%
}

\title{Slip detection for grasp stabilisation with a multi-fingered tactile robot hand}
\author{Jasper W. James, Nathan F. Lepora
\thanks{Manuscript received December 24, 2019; revised August 3, 2020; accepted September 26, 2020. JJ was supported by the EPSRC CDT in Future Autonomous and Robotic Systems (FARSCOPE). NL was supported by a Leadership Award from the Leverhulme Trust on `A biomimetic forebrain for robot touch' (RL-2016-39).}%
\thanks{The authors are with the Dept of Engineering Mathematics, University of Bristol, UK and Bristol Robotics Laboratory, University of the West of England, UK. Email: \{jj16883, n.lepora\}@bristol.ac.uk.}}

\begin{document}
\maketitle
\copyrightnotice
\begin{abstract}
    Tactile sensing is used by humans when grasping to prevent us dropping objects. One key facet of tactile sensing is slip detection, which allows a gripper to know when a grasp is failing and take action to prevent an object being dropped. This study demonstrates the slip detection capabilities of the recently developed Tactile Model O (T-MO) by using support vector machines to detect slip and test multiple slip scenarios including responding to the onset of slip in real time with eleven different objects in various grasps. We demonstrate the benefits of slip detection in grasping by testing two real-world scenarios: adding weight to destabilise a grasp and using slip detection to lift up objects at the first attempt. The T-MO is able to detect when an object is slipping, react to stabilise the grasp and be deployed in real-world scenarios. This shows the T-MO is a suitable platform for autonomous grasping by using reliable slip detection to ensure a stable grasp in unstructured environments. Supplementary video: https://youtu.be/wOwFHaiHuKY
\end{abstract}
\section{Introduction}
Tactile sensing is used by humans when grasping to prevent dropping objects \cite{johansson2009coding}. It is well known that tactile sensing is useful in robotic grasping and provides information complementary to vision \cite{bicchi1989augmentation,luo2015localizing}, such as contact forces and surface texture. With a sophisticated tactile system a robotic hand could replicate many of the complex tasks that humans perform routinely with minimal conscious thought.

One key facet of tactile sensing is slip detection. Early slip detection studies, such as that by Howe and Cutkosky (1989), discussed the importance of slip detection to manipulation whilst presenting a solution using accelerometers embedded in an artificial skin \cite{howe1989sensing}. \textcolor{black}{Scaling up from a single sensor to multiple sensors on a robotic hand presents difficulties such as how to combine control of a many degree-of-freedom (DoF) hand with sensor feedback. Our approach here is to consider the combination of a tactile-enabled manipulator that is midway in control complexity (the 4-DoF Tactile Model-O) and a simple slip-detection method that extends naturally from one tactile fingertip to a many-fingered hand.}

The field of slip detection is diverse and growing with many new sensors and control methods proposed to aid its implementation in grasping and manipulation \cite{Kappassov2015TactileReview,Chen2018}. This progress is based on different technologies, including optical \cite{Khamis2019ASensor,James2018,dongimproved}, neuromorphic \cite{rigi2018novel,Hays2019NeuromorphicControl} and force sensing \cite{stachowsky2016slip, Meier2016TactileDetection} to name but a few. \textcolor{black}{The methods of detecting slip are also diverse with research into both model-based and model-free approaches. Model-based approaches have included use of beam bundle models\cite{VanAnhHo2013BeamMechanism} and friction cones \cite{Reinecke2014ExperimentalSystem}; model-free approaches have included techniques such as random forests \cite{veiga2015stabilizing,Spiers2016Single-GraspSensors} and, recently, deep learning \cite{Meier2016TactileDetection,Li2018SlipInformation, Zhang2018FingerVisionNetwork, Zapata-Impata2019LearningDetection}. In a recent study, Rosset et al. (2019) compared a ConvNet (model-free) to a model using sensor firing rates in a pick-and-place task, with the latter performing better~\cite{Rosset2019ExperimentalDetection}. Clearly, though, the debate on model-based vs model-free is far from being settled, and the merits of both approaches should be investigated.} 
%%%%%%%%%%%%%%%%%%%%%%%%
\begin{figure}[t]
  \centering
    \begin{subfigure}[t]{0.22\textwidth}
        \includegraphics[width=\textwidth]{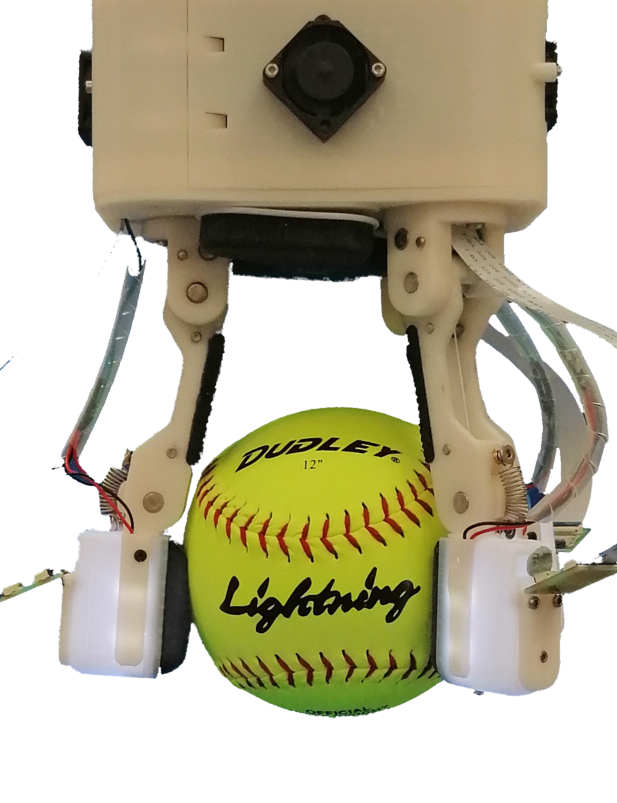}
        \label{fig:ball_up}
    \end{subfigure}
    \begin{subfigure}[t]{0.22\textwidth}
        \includegraphics[width=\textwidth]{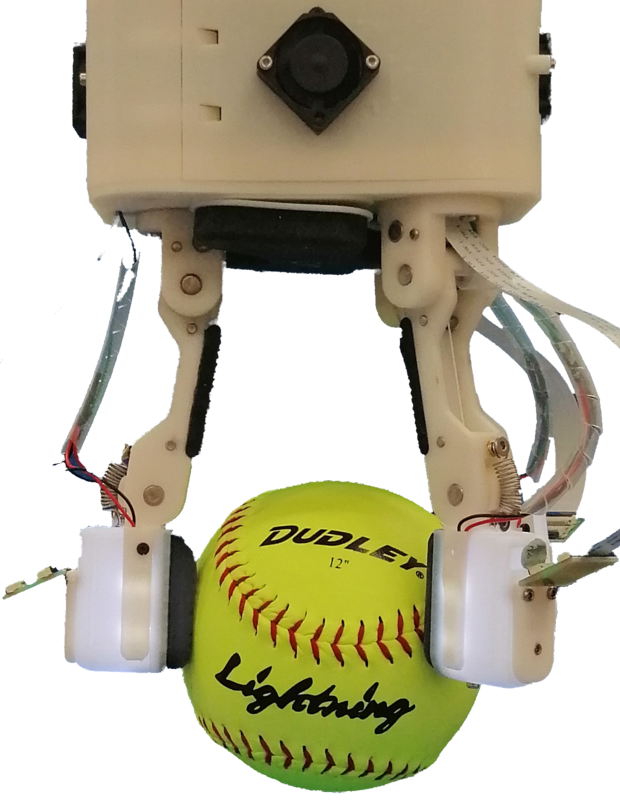}
        \label{fig:ball_down}
    \end{subfigure}
        \caption{The Tactile Model O (T-MO) before and after catching a softball by detecting slip and reacting to prevent it from being dropped.}
        \label{fig:firstpage}
\end{figure}

The aim of this study is to demonstrate the slip detection and correction capabilities of the recently-developed Tactile Model O (T-MO) hand \cite{Church2019} using simple, interpretable methods. This system integrates the TacTip, an optical biomimetic tactile sensor \cite{chorley2009development,ward2018tactip}, onto a GRAB Lab Model O hand \cite{odhner2014compliant}. \textcolor{black}{The principal contributions of this work are to:
\begin{enumerate}
    \item Compare and understand simple slip detection methods based on pin velocity features, which can be scaled from one to three fingers of the Tactile Model-O.
    \item Test multiple slip scenarios, including responding to the onset of slip in real time with eleven different objects in various antipodal pinch grasps (e.g. Fig. \ref{fig:firstpage})
    \item Demonstrate the benefits of slip detection by testing two real-world scenarios: adding weight to destabilise a grasp and \textcolor{black}{repeat detection of slip in real-time to perform first time grasping}.
\end{enumerate}}
\noindent Our conclusion is that the Tactile Model O (T-MO) is able to detect when an object is slipping, react to stabilise the grasp, and be deployed several distinct grasping scenarios. This also demonstrates that the ability to detect slip serves as a useful and robust metric for determining grasp stability in both laboratory and real-world test scenarios.

The paper is laid out as follows. We first give an overview of relevant studies in this field. We then describe the hardware and software methods used before moving onto the experiments. The experiments undertaken in this paper are split into three distinct sections. Firstly, we test the slip detection capabilities of a single sensor using the slip detection rig previously described. Secondly, we test the capabilities of the tactile hand with several natural objects. Finally, we deploy the slip detectors developed during the second test in two applications of slip detection that aid grasping performance.

\section{Background}

%%%%%%%%%%%%%%%%%%%%%%%%%%
\begin{figure*}[t]
\centering
\includegraphics[width=\textwidth]{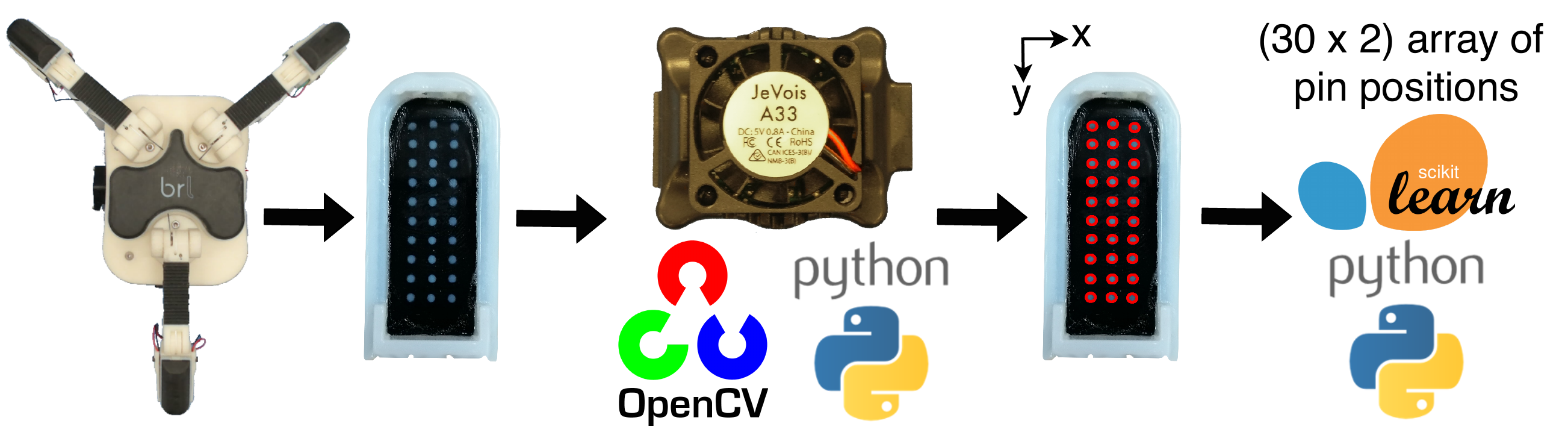}
\caption{Data flow schematic of the Tactile Model O (T-MO). From left to right: T-MO with embedded cameras; TacTip sensor with 30 pins; JeVois vision system captures image of sensor; TacTip pins detected using Python and OpenCV (camera axes shown); (30x2) array of pin coordinates sent to control PC.}
\label{fig:flow}
\end{figure*}
%%%%%%%%%%%%%%%%%%%%%%%%%

The aim of this study is to demonstrate the real-time slip detection and response capabilities of a multi-fingered tactile robotic hand: the Tactile Model-O. James et al. (2018) have previously shown that using a Support Vector Machine (SVM) resulted in effective slip detection capabilities when using an optical tactile sensor, the TacTip \cite{James2018}, mounted on a robot arm. SVMs were also successful when reacting to the onset of slip to prevent several differently shaped objects from being dropped. A further study by James and Lepora (2018) showed that the method generalised to novel common objects of various shapes and compliances \cite{james2018slip2}.

A large proportion of prior research involving slip detection on objects has used 1 or 2 sensors to press an object against a wall or hold an object in a two-fingered pinch grasp. Veiga et al. (2015) used the former with a BioTac tactile sensor \cite{veiga2015stabilizing}; using a Random Forest classifier they achieved up to 100\% success on multiple objects when detecting a slipping object and preventing it from falling. Dong et al. (2017) integrated two GelSight sensors onto a two-fingered gripper and used slip detection to determine whether a grasp was stable and regrasp if necessary \cite{dongimproved}. Subsequently, Li et al. (2018) used visual and tactile information to evaluate the stability of a grasp. With the same hardware they obtained 88\% accuracy compared to 82\% when using tactile feedback alone \cite{Li2018SlipInformation}. 

Dong et al. (2019) used a modified GelSight sensor, the GelSlim, to detect slip and to perform a task: screwing on a bottle cap \cite{Dong2019}. The GelSlim uses a mirror array to reduce the size of the pinch gripper whilst utilising the same sensing technology as the GelSight. By deforming around objects a high resolution image of an object's morphology can be obtained as well as using markers to determine how the object is slipping. A related sensor, the FingerVision, also uses markers embedded in a soft gel to perform slip detection \cite{Zhang2018FingerVisionNetwork}. They use a convolutional LSTM neural network to detect slip and obtain a 97.62\% classification success rate, but do not detect slip in online grasping experiments.

Slip detection research can be broadly divided into three categories: \textbf{gross} slip (investigated here), where the entire sensor surface is slipping; \textbf{incipient} slip, where part but not all of the object is slipping; and slip \textbf{prediction}, where extracted tactile features can predict when slip is about to occur \cite{Chen2018}. Dong et al. (2018) were able to detect incipient slip using the GelSlim sensor to maintain failing grasps \cite{Dong2019}. Su et al. (2015) \cite{su2015force} were able to predict slip was going to occur 30ms before it was detected by an inertial measurement unit atop an object.

Other work has involved increasing the number of sensors and the variety of slip scenarios. Stachowsky et al. (2016) used contact force sensors to detect slip on a held object and also to determine in real time the required grasp strength when picking up various objects~\cite{stachowsky2016slip}. Li et al. (2017) used a Barrett three-fingered hand with a 3x8 tactile array and a force sensor on each fingertip to detect slip. They used an SVM with up to 96\% success rate of detecting slip on a variety of objects, although the hand did not react to prevent the object from being dropped once slip was detected~\cite{li2017learning}. Veiga et al. (2018) used up to five BioTac sensors on two different hands to stabilise grasped objects and prevent them from being dropped \cite{veiga2018hand}. The control mechanism increased normal force on an object when slip was predicted and slowly released its grip to find the optimum grasping strength. Li et al. (2014) also used BioTac sensors to predict when a grasp is failing and deploy two strategies to find a stable grasp \cite{Li2014LearningSensing}. For further discussion of slip detection we direct the reader to a recent review by Chen et al. (2018) \cite{Chen2018}.

Prior work with the TacTip has included integration onto existing robotic hands. Ward-Cherrier et al. (2016) used an M2 gripper, a two-fingered, two-DoF hand, also from the OpenHand project \cite{ma2016m}, and rolled cylinders across the surface using only tactile feedback to test the dexterity of the M2 \cite{ward2016tactile}. Subsequently, Ward-Cherrier et al. integrated the TacTip onto a GR2, two-fingered gripper \cite{rojas2016gr2}, and  rolled cylinders along a trajectory in the hand's workspace over the TacTip's surface. \cite{ward2017gr2}. Recently James et al. (2020) integrated three TacTips onto a modified GRAB Lab Model O \cite{odhner2014compliant} and demonstrated it maintained grasping performance with the addition of tactile sensing to perform tactile object recognition and grasp success prediction \cite{Church2019}; this hand is also used in the present study.

\section{Hardware Description}
\subsection{Tactile Sensor} \label{sec:sensor}
The tactile sensor used in this study is the TacTip \cite{ward2018tactip, chorley2009development} which originally consisted of a flexible moulded urethane hemispherical skin filled with a clear, compliant, polymer blend and sealed with a clear acrylic lens. This design allows the sensor to conform to object shape and then elastically regain its shape when removed from the stimulus.   

A set of pins are arranged on the inside of the sensor surface, with a camera mounted above the sensor to view the pins as the sensor surface deforms. Each captured frame is processed using Python OpenCV to extract the pin positions using the function {\bf SimpleBlobDetector}. \textcolor{black}{In this work, we additionally utilise  processing capabilities within the hand (the JeVois' processors; see below) to extract the pin positions, thereby reducing the computational load on the control PC.}

Recent upgrades to the TacTip keep the same functional principles but utilise 3D printing to facilitate  rapid prototyping and manufacture of new designs. The TacTip surface now comprises a rubber-like 3D-printed surface made from Tango Black+ supported by a rigid cylindrical base made from Verro White. The sensor is filled with silicone (RTV27905) and sealed with a clear acrylic lens. For more details on of this family of tactile sensors, see Ward-Cherrier et al. (2018) \cite{ward2018tactip}.

\subsection{Robotic Hand} \label{sec:robo_hand}
For this study, we use the Tactile Model O (T-MO) developed by James et al. (2019) \cite{Church2019}; this hand is a GRAB Lab Model O \cite{odhner2014compliant} that has been modified to house three TacTip tactile sensors as fingertips (Fig. \ref{fig:flow}). To be suitable as fingertips, the sensors are far smaller than the original TacTip probe, but nevertheless retain the same working principle.

The hand contains three JeVois machine vision systems \cite{Itti2019JeVoisCamera} which are used to capture images at 60 frames per second (340$\times$280 resolution) and track the pin positions (Fig. \ref{fig:flow}). The pin data is then sent to a control PC for input into a classifier and control system. The ability to perform on-board processing reduces the computational load on the control PC, which allows the T-MO to be used robustly at high data sampling rates. This also opens up the possibility of using the T-MO on a fully autonomous mobile system where computational power is at a premium.

The three under-actuated fingers each contain two joints but a single actuatable degree of freedom. This allows the fingers to conform to the shape of any object without needing a complex control framework to individually actuate each joint. Two of the fingers can also rotate through $90\degree$ about their base to allow for different grasping modes. This rotation is mechanically coupled so constitutes only a single degree of freedom. Overall, the T-MO retains much of the grasping capabilities of the original Model O whilst gaining a tactile sense. \textcolor{black}{For a detailed description of the T-MO, we refer to James et al. (2019), both on the fabrication and construction of the hand, and its capabilities in reliably classifying objects and predicting grasp success based on purely tactile data \cite{Church2019}.}

\subsection{Robotic Platform}
The T-MO is mounted on a Universal Robots UR5 six degree-of-freedom robotic arm. The experiments mainly involve the T-MO grasping objects from the YCB Object Set \cite{Calli2015} to test slip detection on natural objects. 

Prior to using the entire hand, it is important to assess the slip detection capabilities of the newly-developed small tactile sensor compared with the `traditional' larger TacTip which had been examined previously by James et al. (2018)~\cite{James2018}. 

Therefore, we used the same slip detection rig, which consists of a 3D-printed shape mounted vertically on a low-friction rail system (Fig. \ref{fig:slip_rig}). The two rails are Igus Drylin N linear guides, which prevent the object from rotating and falling off the rails whilst minimising friction. The purpose of the low-friction rail system is two-fold. Firstly, it provides a very controlled environment to test how well a tactile sensor performs slip detection. Secondly, it allows for direct comparison between different sensors using the same platform. 

\section{Software Methods}
\subsection{Tactile Data} \label{sec:tactile_data}
One of our main goals is to develop and test a classifier that can accurately distinguish between when an object slips over the sensor surface from when it is static (i.e. is held securely). The first step is to process the raw tactile data, which is a 2D list of the pin positions (hardware section \ref{sec:robo_hand}). Our use of pin positions, as opposed to images \cite{Church2019}, is to give a direct measure of slip from the movement of fixed points on the sensor surface. In our previous work, these were shown to be an effective feature set for slip detection~\cite{James2018}. 

An important pre-processing step is to convert the pin positions into velocities, which is implemented by subtracting the previous pin positions from the current positions. \textcolor{black}{As the data is received at 60 Hz this gives the rate of change of pin positions per frame, which is analogous to velocity}. The vector of pin velocities is collected in Cartesian coordinates \((\Delta x_{i},\Delta y_{i})\) for pins \(1\leq i\leq 30\), and transformed to polar coordinates \(\Delta r_{i} = \sqrt{\Delta x_{i}^{2} + \Delta y_{i}^{2}}\) and \(\Delta \theta_{i} = \arctan(\Delta y_{i}/\Delta x_{i})\). To avoid potential zero division errors for $\Delta x_{i}=0$ a small value $\epsilon \lll \Delta x_{i}$ can be added, however in practice, noise in the image and pin detection means this has never occurred. The angular component ($\Delta \theta_{i}$) of the pin velocities is then shifted to have a mean of zero with respect to the $x$-axis of the camera, \textcolor{black}{which means that the data is independent of slipping angle and should thus allow the classifier to detect slip in any direction}. This yields a 60-dimensional time-series output $(\Delta r_{i},\Delta\theta_{i})$. 

\subsection{Classifier Description}\label{sec:classifier_description}
\textcolor{black}{As described in the introduction, our approach here is to use a simple, interpretable slip-detection method that extends naturally from one tactile fingertip to a many-fingered hand. In prior work (James et al. 2018), we showed that a support vector machine is an effective detector of slip on tactile data preprocessed into pin positions for a single sensor~\cite{James2018}. However, that demonstration left unanswered questions such as why the SVM is an appropriate method or whether other classification methods would be better suited.}

Here we justify the appropriate method by comparing three distinct binary classifiers that can detect slip: \textcolor{black}{ first, a threshold on the magnitude of mean pin velocity;} second, a support vector machine applied to either linear or nonlinear combinations of pin velocities; and third a logistic regression (LogReg) method applied to individual pin velocities. These classifiers were chosen to progressively increase in their sophistication while maintaining an interpretation in terms of the pin velocities being a direct measure of slip occurrence.

\subsubsection{\textcolor{black}{Threshold Classifier}}
\textcolor{black}{This first decision method simply takes the magnitude of the average pin velocities for each time sample and uses a threshold to decide whether slip is occurring or not. Since the pin speeds are expected to be higher and the velocities co-linear when slip occurs, data falling above the threshold $T$ is considered to be slipping. The vector of thirty pin velocities is reduced to a single dimension and classified as slipping if it exceeds a threshold $T$:
\begin{equation}
v=\sqrt{\left(\frac{1}{30}\sum_{n=1}^{30}\Delta x_{i}\right)^{2} + \left(\frac{1}{30}\sum_{n=1}^{30} \Delta y_{i}\right)^{2}}\geq T.
\end{equation}
This metric means that frames where the pin velocities have a wide angular distribution will give a smaller value than when they are co-linear. Intuitively, one might expect that this method would be effective at detecting slip with the TacTip pin velocities because these directly indicate slip. However, as we will see in the results, this classifier suffers from not rejecting false positives; i.e. it does not distinguish slip from changes of pin position unrelated to slip.}

\subsubsection{Support Vector Machine}
SVMs work by separating multi-dimensional data into two classes, using a hyperplane that best separates the two sets of training data. In practise, the decision boundary is constructed from finding two parallel hyperplanes that maximally separate the classes, then taking the hyperplane midway between. If the data cannot be linearly separated, it can be transformed using a nonlinear kernel before constructing the hyperplanes. Here we test both a linear and a Gaussian-kernel SVM to find the best classifier. \textcolor{black}{For the pin velocity data considered here, the SVM classifiers can be interpreted as detecting slip by where the vector of pin velocities falls relative to the hyperplane.}
%%%%%%%%%%%%%%%%%%%%%%%%
\begin{figure}[t]
  \centering
  \text{Single Finger Slip Detection Rig}
    \begin{subfigure}[t]{0.24\textwidth}
        \begin{overpic}[width=\textwidth]{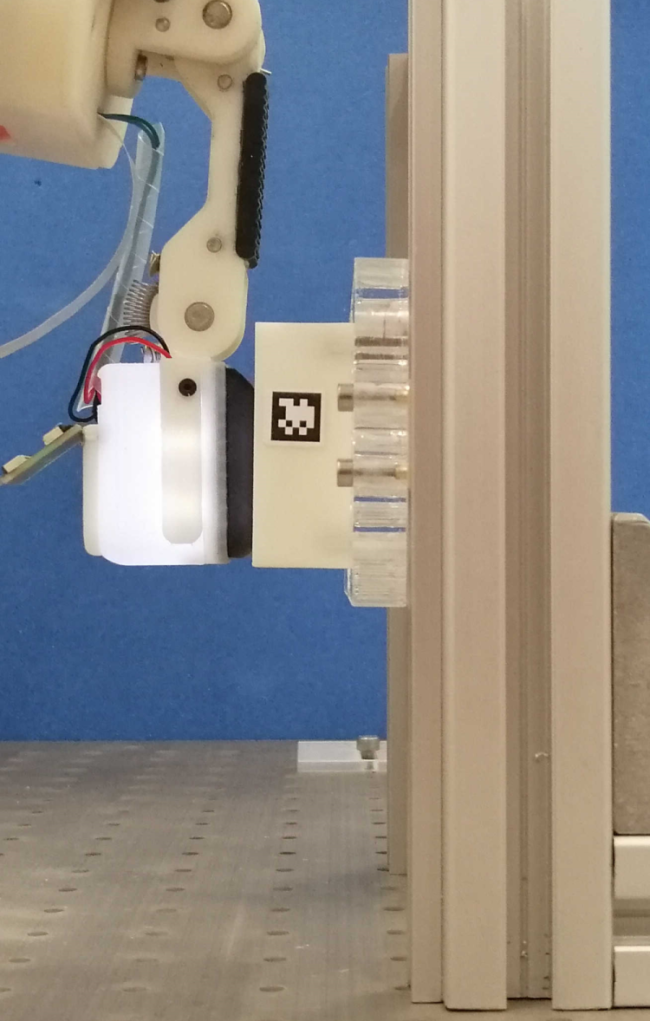}
        \put(3,9){\large Low friction}
        \put(12,4){\large rail}
        \put(17,14){\linethickness{0.8mm}\vector(1,1){20}}
        \put(34,68){\linethickness{0.3mm}\dashline{4}(0,0)(25,0)}
        \end{overpic}
        \label{fig:sliprig_up}
    \end{subfigure}
    \begin{subfigure}[t]{0.24\textwidth}
        \begin{overpic}[width=\textwidth]{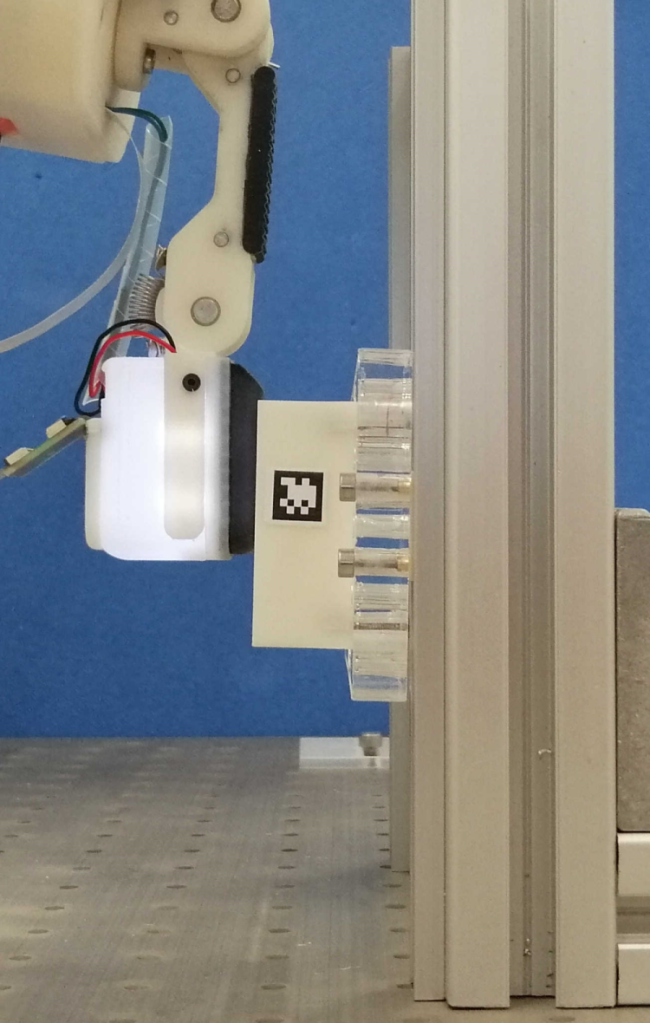}
        \put(19,9){\large ArUco}
        \put(18.5,4){\large Marker}
        \put(29,14){\linethickness{0.8mm}\vector(0,1){32}}
        \put(2,67.8){\linethickness{0.3mm}\dashline{4}(0,0)(40,0)}
        \end{overpic}
        \label{fig:sliprig_down}
    \end{subfigure}
    \caption{Low friction rail system for testing slip detection capabilities of a single sensor. (L) Flat object being held prior to slip occurring. (R) Object being caught after slip has been detected; a clear drop in height is present which is detectable by the ArUco optical marker.}
    \label{fig:slip_rig}
\end{figure}
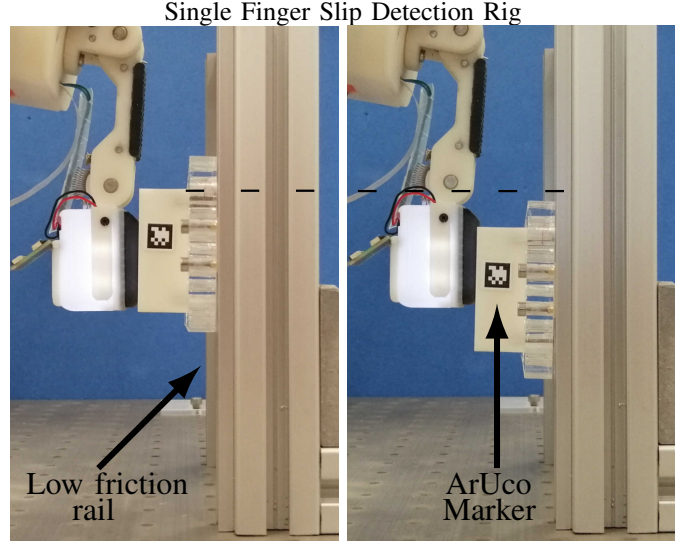  
%%%%%%%%%%%%%%%%%%%%%%%%%

For the case of a Gaussian kernel, two additional hyperparameters must be optimised. The first is a regularization parameter $C$ that gives the trade off between correctly classifying data and maximally separating the two classes: a small $C$ results in a narrow boundary (distance between hyperplanes) but many misclassifications, and vice versa for large $C$. The second hyperparameter, $\gamma$, is the kernel scale, which can be seen as setting the complexity of the class boundary. For small values, $\gamma$, will have little effect on the data and the decision boundary will be similar to a linear SVM; for large values the decision boundary can be very complex, by fluctuating to place individual data points on either side of the boundary, and therefore has a tendency to overfit. For a more detailed description on SVMs see \cite{Cortes1995Support-vectorNetworks}.

\subsubsection{Logistic Regression}
\textcolor{black}{LogRegs are an effective machine learning technique that because they are simple to implement are commonly used to provide a baseline when comparing to other more complex methods~\cite{Jurafsky2009SpeechRecognition}}. They are binary classifiers which assign an observation to a class given a set of input variables, in this case pin velocities, by using the logistic function to give a hypothesis based on a decision boundary 
\begin{equation}
    h_{\theta}(\bm{x}) = \frac{1}{1+e^{-\bm{\theta} \cdot \bm{x}}},
\end{equation}
where $\bm{x}=(1,x_{1},x_{2},...,x_{m})$ is a vector of observations and $\bm{\theta}=(\theta_{0},\theta_{1},\theta_{2},...,\theta_{m})$ are the model parameters to be determined from optimizing the fit of $h_\theta$ to the class labels. In our case, the prediction will be class 1 (slip) if
\begin{equation}
    \theta_{0} + \theta_{1}x_{1} + ... + \theta_{m}x_{m} \geq 0
\end{equation}
\textcolor{black}{As here these $m$ observables are the pin velocities, the LogReg method may be considered as analogous to the threshold classifier above but instead using individual pin velocities.}

To avoid overfitting, we introduce a regularization parameter $C$ that penalises large fluctuations in the model parameters. All the SVMs and LogRegs here are trained using Python's scikit-learn package with functions \textbf{SVC} and \textbf{LogisticRegression} respectively. Hyperparameters are found using Bayesian optimization with the scikit-optimize function \textbf{BayesSearchCV}.

%%%%%%%%%%%%%%%%%%%%%%%%
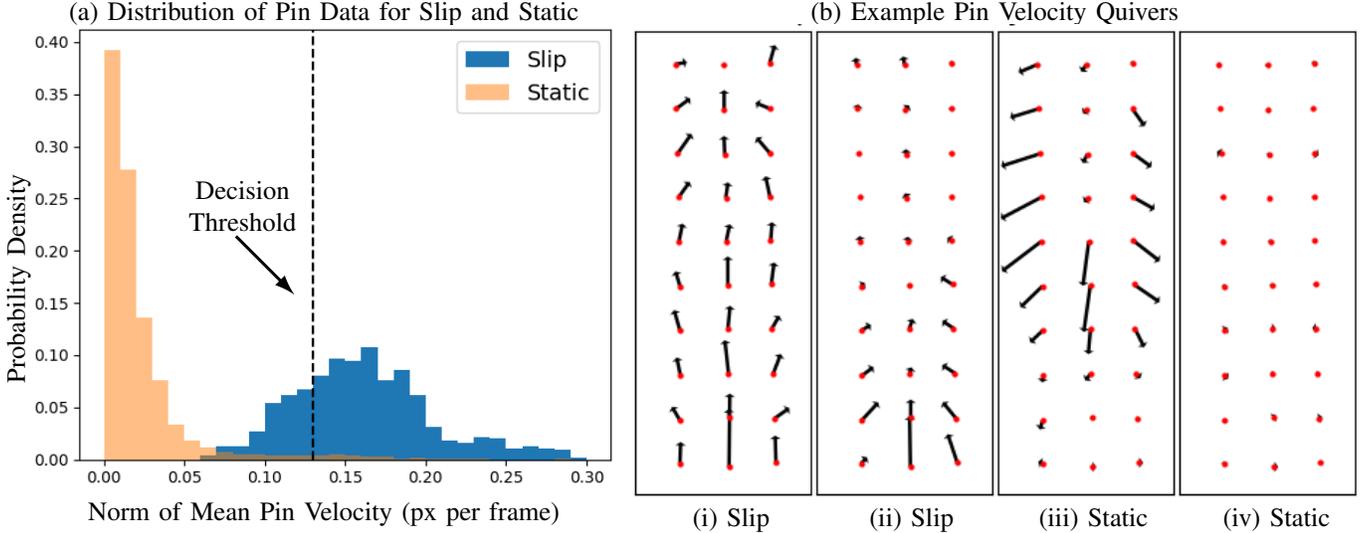
\begin{figure*}[t]
\textcolor{black}{
  \centering
    \hspace{3mm}
    \begin{subfigure}[t]{0.43\textwidth}
        \centering
        \text{(a) Distribution of Pin Data for Slip and Static}
        \begin{overpic}[width=\textwidth]{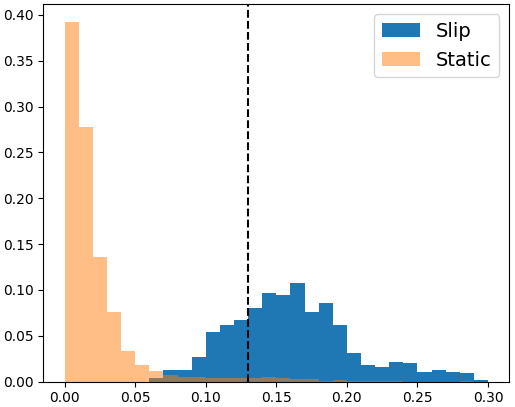}
        \put(-4,18){\rotatebox{90}{Probability Density}}
        \put(27,47){\begin{varwidth}{8cm}\centering Decision\\Threshold\end{varwidth}}
        \put(35,43){\linethickness{0.4mm}\vector(1,-1){10}}
        \end{overpic}
        \text{Norm of Mean Pin Velocity (px per frame)}
    \end{subfigure}
        \centering
    \begin{subfigure}[t]{0.54\textwidth}
        \centering
        \text{(b) Example Pin Velocity Quivers}
        \begin{overpic}[width=\textwidth]{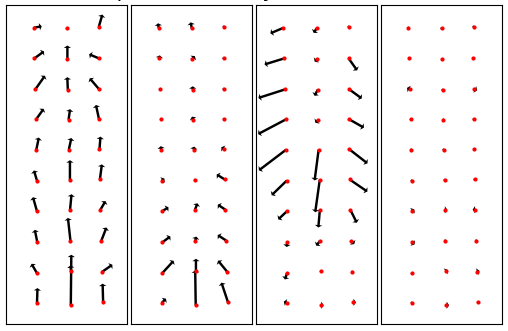}
        \put(9,-3){(i) Slip}
        \put(33,-3){(ii) Slip}
        \put(56,-3){(iii) Static}
        \put(81,-3){(iv) Static}
        \end{overpic}
    \end{subfigure}}
    \caption{\textcolor{black}{(a) Probability density of magnitude of mean pin velocity for all slip and static samples in the training set with optimal decision threshold ($T$) marked. The static distribution has a long tail which is what causes the threshold to be unreliable. The long tail is barely visible due to the large imbalance in sample numbers therefore if the threshold were moved into the valley between distributions there would be an unacceptable number of false positives. (b) Example pin velocities: (i) slip - right of $T$;  (ii) slip - left of $T$; (iii) static - right of $T$; (iv) static - left of $T$. (iii) has large arrows with a sufficiently large mean vertical velocity component to place it on the wrong side of the boundary leading to an incorrect classification. (ii) is a sample from several frames after slip onset where the velocities have shrunk resulting in a value $<T$ and incorrect classification.}}
    \label{fig:hist}
\end{figure*}
%%%%%%%%%%%%%%%%%%%%%%%%%
\subsection{Classifier Evaluation}
\textcolor{black}{A subtlety in applying the above classifiers to slip/non-slip data is that there are usually more examples of non-slip than slip, but we do not want to discard data that could give a better classifier.} Then the normal method of using accuracy to evaluate the classifier ceases to be a good method. A more reliable metric in such cases is the F1 score, which takes into account the Precision and Recall defined as
\begin{subequations}
\begin{tabularx}{0.5\textwidth}{Xp{0.5cm}X}
\begin{equation}
    P = \frac{TP}{TP+FP}
\end{equation}
& &
\begin{equation}
    R = \frac{TP}{TP+FN}
\end{equation}
\end{tabularx}
\end{subequations}
where TP is the true positive rate, FP is the false positive rate and FN is the false negative rate. The F1-score is then:
\begin{equation}
    F1 = 2\frac{P\times R}{P+R}.
\end{equation}
As this does not take into account true negatives we calculate the F1 score for each class and average the result.
%%%%%%%%%%%%%%%%%%%%%%%%
\begin{figure*}[t]
  \centering
    \begin{subfigure}[t]{0.405\textwidth}
        \includegraphics[width=\textwidth,trim={0.5cm 0cm 1.2cm 0.5cm},clip=true]{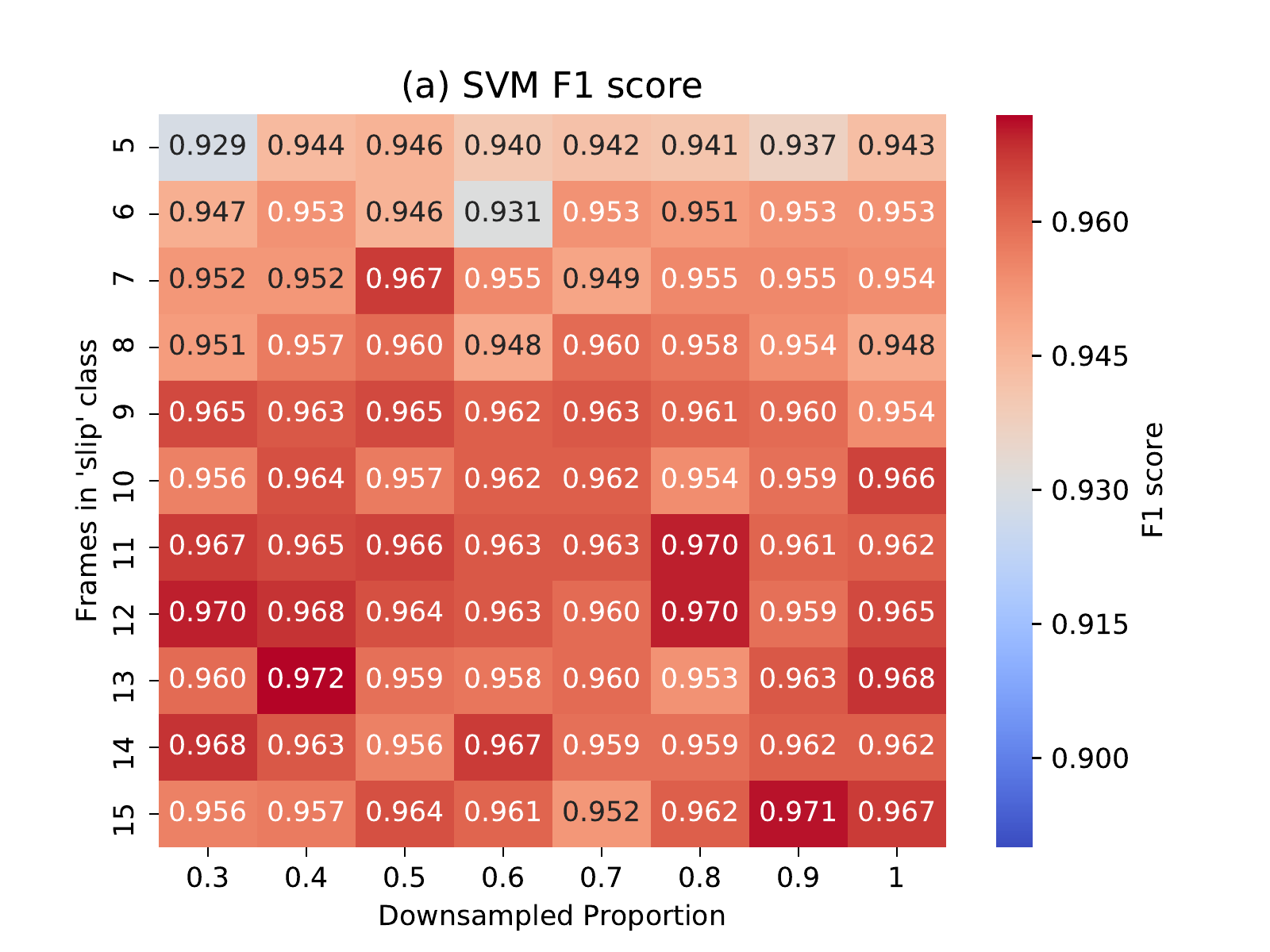}
    \end{subfigure}
        \centering
    \begin{subfigure}[t]{0.405\textwidth}
        \includegraphics[width=\textwidth,trim={0.5cm 0cm 1.2cm 0.5cm},clip=true]{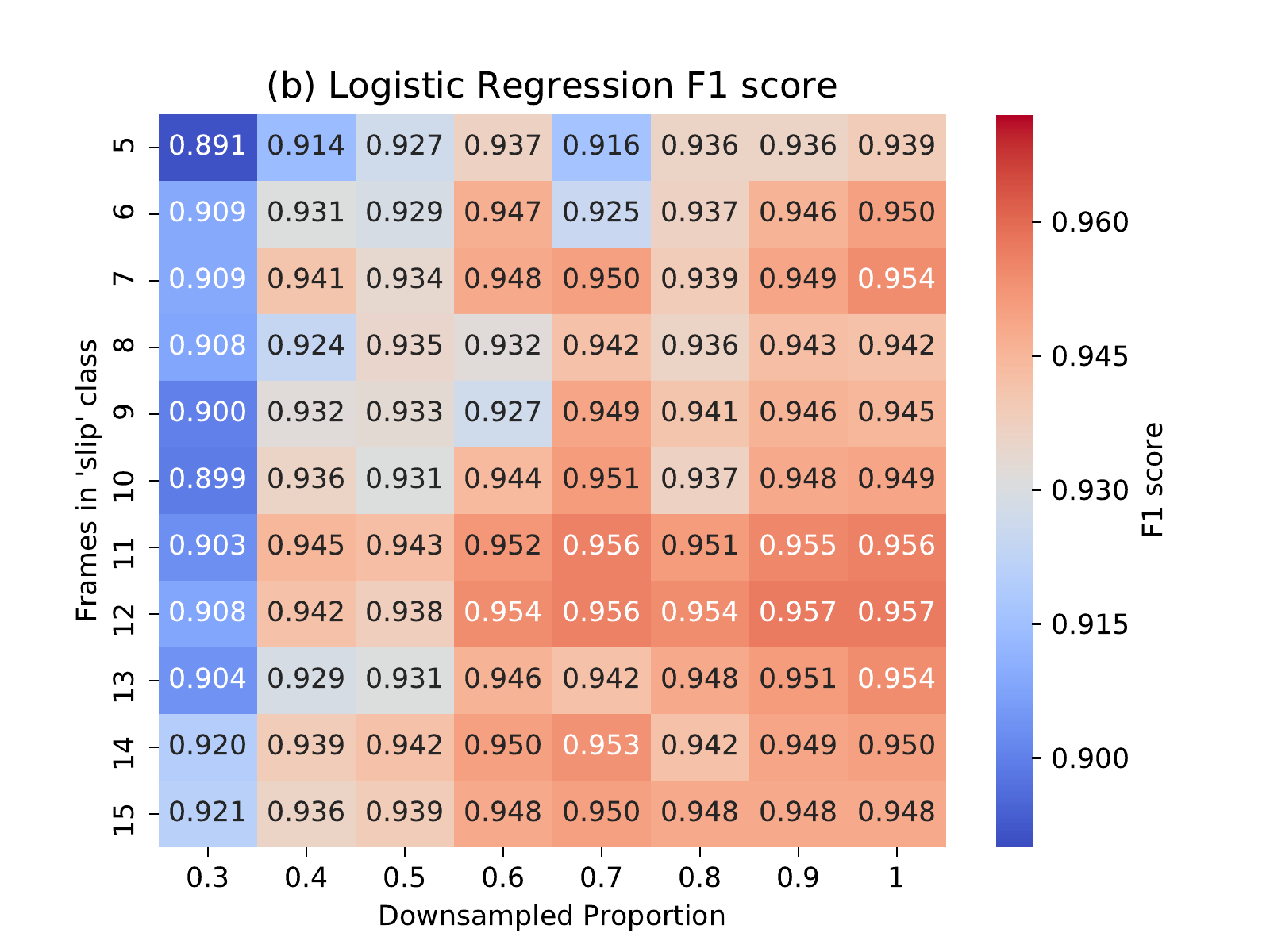}
    \end{subfigure}
    \begin{subfigure}[t]{0.175\textwidth}
        \includegraphics[width=\textwidth]{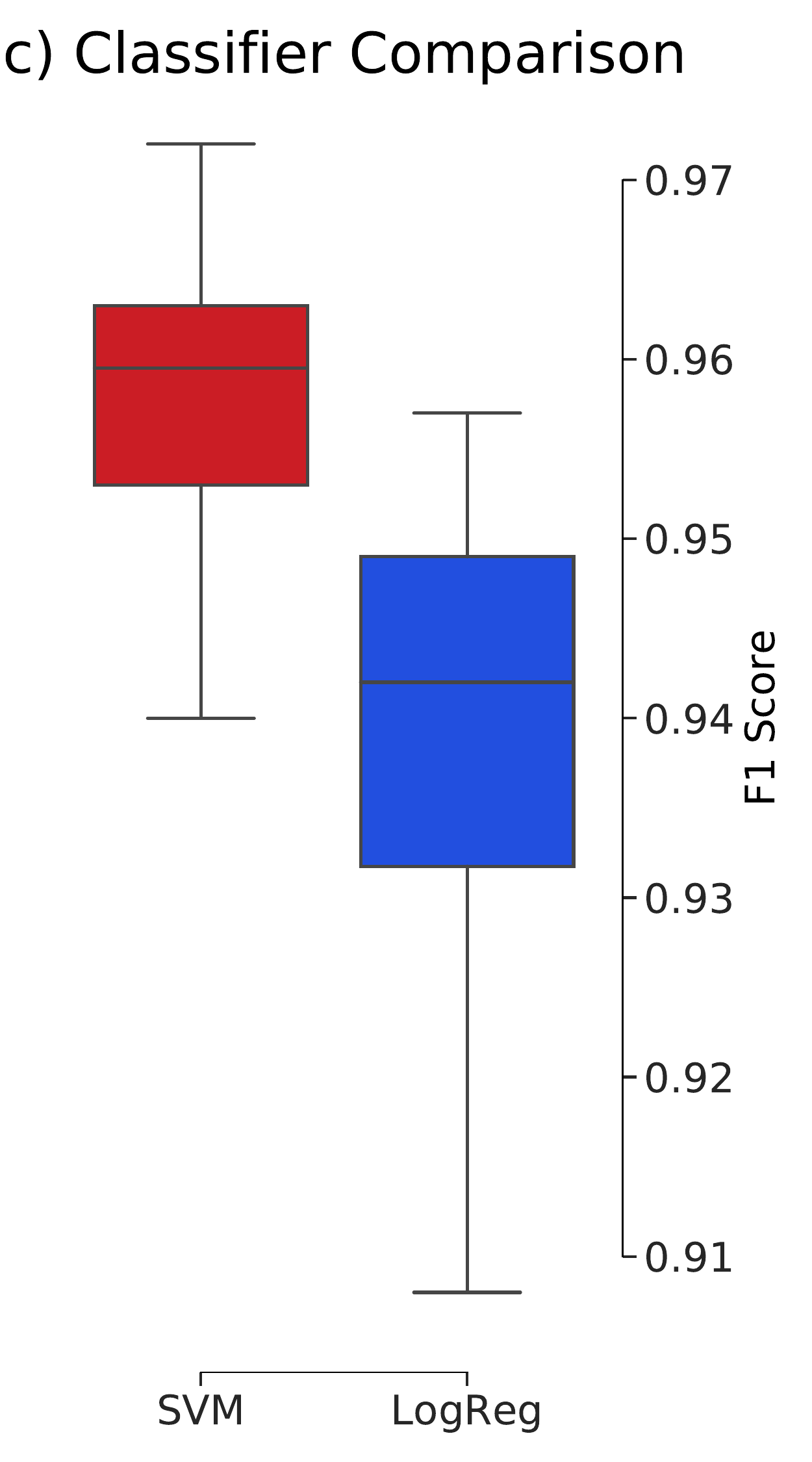}
    \end{subfigure}
    \caption{Heatmaps of F1 scores of (a) Logistic regression and (b) SVM when altering the amount of data in the `static` class by randomly downsampling and altering the number of frames used in the `slip' class. (c) Boxplot showing direct comparison of the two methods. The SVM is not significantly affected by downsampling and consistently outscores the logistic regression. Both methods trail off significantly when downsampling below 30\% and with fewer than 5 frames in the `slip' class; these results are omitted for clarity.}
    \label{fig:lr_svm}
\end{figure*}  
%%%%%%%%%%%%%%%%%%%%%%%%%
\section{Single Finger} \label{single_finger_training}
\subsection{Experiment Description}
The first task uses the rail-system (Fig. \ref{fig:slip_rig}) to test the slip detection capabilities of a single tactile fingertip of the T-MO. The rotationally-fixed finger (bottom finger of left hand side of Fig. \ref{fig:flow}) held the 3D printed object at the top of the rail system by pressing against it and then retracting at a random speed between 0.1 and 5 mm s$^{-1}$ to induce a slip. The tactile fingertip
stopped recording when the UR5 robot arm had finished retracting and therefore, due to the variation in speed, each test has a different number of frames. For this test, we collected data using only a single finger.

To label the data we placed an ArUco optical marker on the slipping object and recorded its height with another JeVois camera. Any frames from the TacTip that arrive after the ArUco begins to fall are labelled as belonging to the \textbf{slip} class and any that arrive prior to this are labelled as \textbf{static}. This makes the labelling process autonomous and contains no labelling bias that could have occurred if performed manually.

As the amount of time spent not slipping vastly outweighs the time spent slipping, there is a large class imbalance between the \textbf{static} and \textbf{slip} classes. To have enough data in the \textbf{slip} class we ran the data collection procedure 100 times, which will be evenly divided between training and test sets. However, there is another issue. Whilst finding the boundary between \textbf{static} and \textbf{slip} is simple, finding the point at which the slip ends is somewhat arbitrary. We wish to confidently find the boundary between non-slip and slip and then reliably detect the continuation of slip for as long as possible, so that an action can be taken to minimise slippage. Therefore, we varied the number of frames $n_{\rm slip}$ in the \textbf{slip} class from $1$ to $15$ to obtain a classifier that is reliable at detecting slip for a given number of frames after slip commences.

We tested three methods to find the best offline classifier: a threshold on magnitude of average pin velocity, a logistic regression and a support vector machine (Methods Section~\ref{sec:classifier_description}). \textcolor{black}{To summarise, the threshold attempts to identify whether a spiking feature in the aggregate pin speed can be accurately identified as slip, whereas the SVM and LogReg examine each individual pin velocity to draw a complex decision boundary that distinguishes slipping from static touch}.

As is common in many machine learning methods, large datasets can be computationally challenging (e.g. the time taken to optimize an SVM scales as the number of samples squared \cite{Wang2014TrainingTraining}). Therefore, for efficiency, we test the effect of down-sampling this data. To find the best classifier offline, we vary the number of frames in the `slip' class and randomly down-sample the data before optimising each classifier. \textcolor{black}{Down-sampling by a proportion $d$ leaves a dataset of size $N' = d N_{0} + N_{1}$ for a dataset with $N_{0}$ samples in the static class and $N_{1}$ samples in the slip class (note that the slip class is not down-sampled, because the effect of varying the slip samples is already being considered).} The optimised classifiers are then tested on the entire test set.

\subsection{Results}
\subsubsection{Offline performance}\label{sec:single_finger}
The first test was an offline characterisation of slip detection using pre-collected data from a single finger. 

\textcolor{black}{{\bf Threshold method:} The threshold was chosen by testing ten values between the magnitude of mean pin velocity across all static samples and all slipping samples. The best performing threshold was an average of 0.13 pixels/second, which had an $F1$-score of 0.753 on the complete dataset with eleven frames in the `slip' class. As we will see, this is significantly worse than the SVM and LogReg classifiers (that typically have an $F1$s-score above 0.9), and so demonstrates that a more sophisticated classifier is necessary to capture the complexities of the boundary between slipping and static touch.}

\textcolor{black}{The poor performance of the threshold method appears due to the optimal threshold overlapping with much of the slip class (Fig. \ref{fig:hist}a). This is visible in pin velocity quiver plots from the static class (Fig.~\ref{fig:hist}b(iii)) that would be misclassified as slip with the threshold classifier: this data is from an initial contact with an object that is not slip but still results in large pin velocities (although the distribution remains significantly different from slipping). Similarly, another quiver plot shows slip that would be misclassified as static (Fig.~\ref{fig:hist}b(ii)) because it occurs late after slip onset which leads to a weaker slip signal. Overall, the classifier should be robust to such effects and thus this method is not appropriate for slip/static separation.} 

{\bf SVM and LogReg methods:} We tested these two classifiers under the effects of downsampling the training data and varying the number of frames in the slip class. 10\% to 100\% of the data was retained in 10\% increments, varying the number of frames in the slip class from one to fifteen.

The SVM outperformed the logistic regression across all tests (Fig.~\ref{fig:lr_svm}) giving peak $F1$ scores of 0.972 (SVM) and 0.957 (LogReg). The logistic regression showed a noticeable drop in performance when down-sampling the data below 60\%, whereas the SVM performed well until just 30\% of the data was kept (even scoring best when down-sampled to 40\%). The mean $F1$-score from all SVM  classifiers in Fig.~5 is 0.960 which outscores the best score for the LogReg (0.957). Therefore, the SVM is the superior method for detecting slip using the TacTip.  

%%%%%%%%%%%%%%%%%%%%%%%%
\begin{figure*}[t]
  \centering
    \begin{overpic}[width=0.99\textwidth,trim={0cm 0cm 0cm 0.5cm},clip=true]{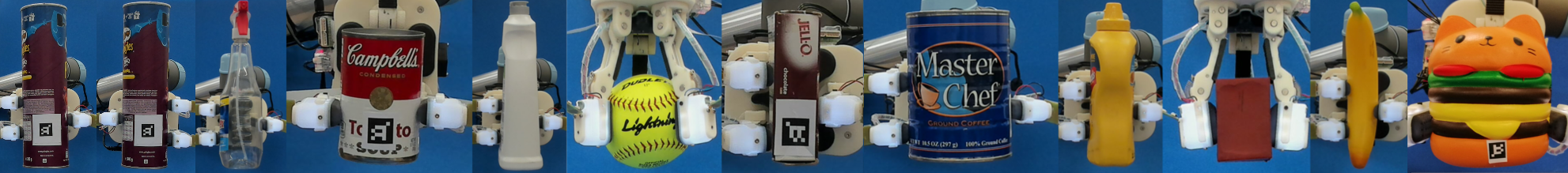}
    \put(0.5,11){\small Can 0\degree\ }
    \put(5.7,11){\small Can 180\degree\ }
    \put(13,11){\small Windex}
    \put(22.5,11){\small \color{red}Soup}
    \put(30,11){\small Bleach}
    \put(40,11){\small \color{red}Ball}
    \put(49,11){\small \color{red}Jello}
    \put(58.5,11){\small \color{red}Coffee}
    \put(68,11){\small Mustard}
    \put(77,11){\small \color{red}Brick}
    \put(84,11){\small Banana}
    \put(94,11){\small \color{red}Cat}
    \end{overpic}
    \caption{Grasps used for online testing with eleven different natural objects. We tested with the hand in various orientations and used different number of fingers in contact with the object to demonstrate robustness. The left two images show the crisps can held at 0\degree\ and 180\degree\ for testing in Section \ref{sec:hand_online_pringles}. Objects in black were used for collecting training data. All objects (including the remaining six novel objects, in red) were used for testing.}
    \label{fig:all_grasps}
    % Make clear which are training and testing - sort out labels!!
\end{figure*}
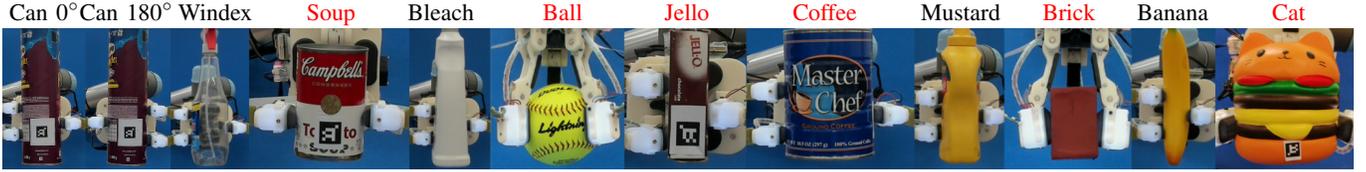  
%%%%%%%%%%%%%%%%%%%%%%%%%
However, classification performance is not the only important factor: a fast reaction time is also critical for catching a slipping object quickly. Therefore, it was also important to ensure both the SVM and LogReg classification methods can classify new data quickly. We verified that both are able to classify in sub-millisecond time, which is much smaller than the response time of the hand (>5ms). This allows us to select the method going forward based purely on score. 

Classifiers using $n_{\rm slip}$ from 11-15 frames performed best for both methods, so it was hard to select with confidence the ideal number of frames to choose for future tests. That said, the best scoring classifier had the following parameters: 

%%%%%%%%%%%%%%%%%%%%%%%%%%%%%%%%
\begin{table}[H]
\centering
\begin{tabular}{cccccc}
\textbf{Classifier} & \textbf{Kernel} & $\gamma$ & $C$ & DS & $n_{\rm slip}$ \\
SVM                 & Gaussian        & 1.104          & 3.989      & 0.4         & 13          
\end{tabular}
\end{table}
%%%%%%%%%%%%%%%%%%%%%%%%%%%%%%%%
%%%%%%%%%%%%%%%%%%%%%%%%%%%%%%%%%
\begin{table}[b]
\centering
\begin{tabular}{|l|l|l|l|l|}
\hline
\textbf{N Frames} & \textbf{Caught (\%)} & \textbf{FP (\%)} & \textbf{Dropped (\%)} & \textbf{d (mm)} \\ \hline
1          & 74               & 20               & 6                  & 21              \\ \hline
2          & 72               & 0                & 28                 & 29              \\ \hline
\end{tabular}
\caption{Online slip detection and catch success. High success at catching the object was achieved in both tests but waiting for two consecutive frames to be classified as `slip' greatly reduces the false positive responses. In the second scenario slip was correctly detected in each test but the single finger was unable to provide enough force to stop the object in time. \textbf{N Frames} refers to the number of consecutive frames classified as slip before an action is taken. \textbf{FP} is false positive (action triggered before slip) and \textbf{d} is the average distance slipped before being successfully caught.}
\label{tab:rig_results}
\end{table}
%%%%%%%%%%%%%%%%%%%%%%%%%%%%%%%%
\noindent where $\gamma$ is the kernel scale, $C$ is the box constraint, DS is the down-sampling rate and $n_{\rm slip}$ is the number of frames used in the slip class per test. These parameters will be used for online testing and to train any future classifiers.

\subsubsection{Online performance}
Having trained the best classifier offline, we now test the ability of the platform to detect slip in real time. We used the same rig and experiment setup as for collecting data, except now the finger and UR5 perform an action to minimise slippage when detected.

Again, the object is held on the low friction rail and the UR5 is retracted. When slip is detected, the UR5 moves 1.5mm further forward and the torque on the finger is increased slightly. \textcolor{black}{These were justified, as once the object starts moving more force is required to stop it than to hold it without slipping; however, moving too far forward risks damaging the sensors by over-compressing them. 1.5mm compression gives sufficient additional force without damaging the sensors}.

Two reaction strategies are used: first, if any frame was classified as slip, the predefined action is performed to catch the object; second, two consecutive frames much be classified as slip before the action is taken. The latter reduces the likelihood of false positives triggering an action too early at the cost of slower response times. We tested each scenario 50 times. An ArUco marker was again attached to the object to track how far the object slipped before being caught and to independently verify whether a test failed or an action was taken before slip occurred. 

These two scenarios gave similar task success (Table \ref{tab:rig_results}). The major difference is in false positive rate which drops from 20\% to 0\% when using two consecutive frames to trigger an action rather than a single frame. 

It should be noted that the second scenario never failed to detect the slip, and the lower catch rate was due to it being unable to stop it in time. We think this is due to the larger amount of force being needed to arrest the slipping object once downwards momentum has built up, and the small size of the object. The finger is also compliant, which gives a small latency between moving the motors and a force being applied. We anticipate that there will be additional subtleties when using all three fingers to grasp an object, rather than using a single finger and a low friction rail, which we now examine.

\section{Whole Hand Testing}
\subsection{Experiment Description} \label{sec:hand_exp_des}
Having determined the single finger performance, we now test the effectiveness of the T-MO hand at performing slip detection and re-grasping natural objects with all three fingers. This gives a good indicator of how well the T-MO will perform when slip detection is deployed in realistic scenarios.

For consistency, we follow a similar approach to the previous section in regards to training data collection and testing. We therefore select objects that can be easily grasped and tracked by an ArUco marker: five objects from the YCB Object Set \cite{Calli2015} that vary in shape, weight and surface smoothness and yet can be reliably tracked to detect the onset of slip for training data labelling (objects with black titles in Fig. \ref{fig:all_grasps}).

The primary difficulty in scaling from one finger to three is that the dynamics of the slip onset become more complex. For example, one finger may begin to slip before another when a grasp is relaxed and, crucially, before the object starts to drop (as registered by the ArUco marker). To combat this, we used a constant finger movement speed when deliberately dropping the object ($\approx0.5\%$ of the maximum \textcolor{black}{finger movement speed; $\sim$0.07 rad s$^{-1}$)} which we found to be a good compromise between moving too slow (which contaminates the dataset labels when some sensors slip before others) and too fast (which results in slip before sufficient tactile data is collected). To find the best classifier for use in real-time scenarios, we trained classifiers under two conditions, as described below. 

First, we trained a distinct classifier for each finger, using training data from that single sensor alone. Each of these three classifiers can then be used to classify new data from the sensor from on which it was trained. The justification is that each manufactured TacTip will have slightly different properties that may lead to a one-size-fits-all classifier being inappropriate. We will refer to these three classifiers collectively as the \textbf{local} classifier. The drawback of training classifiers on a per-sensor basis is that in different grasps the load may be distributed differently among the three sensors. 

The second approach was to train a single classifier on all the data from all sensors to encapsulate a more general picture of what constitutes slip and ultimately be more versatile. We will refer to this classifier as the \textbf{global} classifier. This global classifier has the advantage of having knowledge from multiple grasp loads without the disadvantage of needing collect data in many different orientations.

\textcolor{black}{In both conditions, we deploy a classifier on data from each sensor asynchronously: for the \textbf{local} it is a different classifier for each sensor and for the \textbf{global} it is the same. This is necessary as, at any time, some sensors can be slipping whilst the others are not; also, each sensor's frame rate can vary slightly, so a synchronous method would have longer latency.}

\textcolor{black}{In total we have four classifiers, three for the local case (each finger) and one for global.} The best classifiers, trained offline, in both of the above cases can then be tested with online slip detection and regrasping, using the five objects used for data collection and with another five novel objects from the YCB dataset (red titles in Fig. \ref{fig:all_grasps}). \textcolor{black}{We also use a sixth novel object, a highly deformable cat which we test both on its own and with a 100g mass adhered to the underside.}

As demonstrated in the single finger case, we can change how we react to slip to combat limitations in the classifier such as false positives. With three sensors we increase the parameters at our disposal for developing a strategy to respond to slip, which should be sufficient to balance any drop in classifier performance from scaling up from a single sensor to three. The parameters we vary are 
\begin{enumerate}
    \item Number of consecutive frames detecting slips: N\textsubscript{Fr} 
    \item Number of sensors detecting slip: N\textsubscript{Sen}
\end{enumerate}
The frames variable allows each sensor to use more information to decide a slip classification, reducing false positives. \textcolor{black}{The sensor variable allows a longer wait until more sensors are slipping before the hand responds}. Both variables determine how cautious we should be in making a slip decision. 

\subsection{Results}
\subsubsection{Offline Performance}
For training on the five objects from the YCB object set (Fig. \ref{fig:all_grasps}, black titles), we held each object in a cylindrical pinch grasp (thumb opposed to other two fingers) by closing the fingers until the object was stable. The grasp was then released until the object slipped. This was performed twenty times per object with an equal split between training and testing sets.

Overall, there is a total of $50$ training runs ($10$ per object) and $50$ test runs. The ArUco marker on each object allows us to label each frame as either slipping or static. Motivated by the single-finger results (Section~\ref{sec:single_finger}), we down-sample the number of frames in the \textbf{static} class to $40\%$ of its original size to bring the balance of frames in each class and reduce training times. We use thirteen frames in the \textbf{slip} class, which likewise performed best on a single finger.

In total four classifiers were trained: \textcolor{black}{one for the global classifier (all fingers) and three for the local (each finger)}. As expected, the local classifier performs better with an average F1-score of 0.833, against 0.822 for the global classifier. \textcolor{black}{Note that for comparison we also tested the classifier trained previously on just a single finger using the slip rig (Section~\ref{single_finger_training}): this scored 0.627, which demonstrates that the data is different when grasping an object with the entire hand.}

Considering the three local classifiers individually gives $F1$-scores of 0.869, 0.824, 0.805 respectively. The higher $F1$-score of the first sensor is consistent with it being opposed to the other two sensors: it will be compressed more (to balance the force from two other fingers), making slip more apparent.

Notice also the drop in $F1$-scores from the single finger ($0.869$ maximum) to the whole hand ($0.833$). \textcolor{black}{We attribute this to the increased complexity of the dynamics with an entire hand, which demonstrates the difficulty in scaling up from one sensor to three}. For a single finger, the object either slips or it is static, and so labelling slip is straightforward; however, with three fingers the object can slip on some fingers before the object falls. Thus the labelling can confuse the stick-slip boundary and therefore reduce the success of the classifier. That said, results from an offline should be treated with caution, because the true test is under online conditions of slip detection and regrasping, which we cover next.

%%%%%%%%%%%%%%%%%%%%%%%%%%%%%%%%%%%%%%%%%%%%%
\begin{table}[b]
\centering
\begin{tabular}{|l|l|l|l|l|l|l|l|}
\hline
SVM & Strategy & Grasp   & TP & FP & FN & D (mm)  \\ \hline
Local      & \multirow{2}{*}{$2\textsubscript{Fr} 3\textsubscript{Sen}$}       & 0\degree  & 100 & 0  & 0  & 39 \\ \hhline{-~-----}
Global     &                                         & 0\degree  & 100 & 0  & 0  & 40 \\ \hline
Local      & \multirow{2}{*}{$2\textsubscript{Fr} 2\textsubscript{Sen}$}       & 0\degree  & 100 & 0  & 0  & 25 \\ \hhline{-~-----}
Global     &                                        & 0\degree  & 100 & 0  & 0  & 28 \\ \hline
Local      & \multirow{2}{*}{$1\textsubscript{Fr} 3\textsubscript{Sen}$}       & 0\degree  & 100 & 0  & 0  & 35 \\ \hhline{-~-----}
Global  &                                            & 0\degree  & 100 & 0  & 0  & 35 \\ \hline
Local      & \multirow{2}{*}{$1\textsubscript{Fr} 2\textsubscript{Sen}$}       & 0\degree  & 60  & 40 & 0  & 17 \\ \hhline{-~-----}
Global   &                                            & 0\degree  & 80  & 20 & 0  & 16 \\ \hline
Local      & \multirow{2}{*}{$2\textsubscript{Fr} 3\textsubscript{Sen}$}       & 180\degree & 70  & 0  & 30 & 63 \\ \hhline{-~-----}
Global    &                                          & 180\degree & 100 & 0  & 0  & 38 \\ \hline
Local      & \multirow{2}{*}{$2\textsubscript{Fr} 2\textsubscript{Sen}$}       & 180\degree & 90  & 0  & 10 & 45 \\ \hhline{-~-----}
Global     &                                         & 180\degree & 100 & 0  & 0  & 27 \\ \hline
Local      & \multirow{2}{*}{$1\textsubscript{Fr} 3\textsubscript{Sen}$}       & 180\degree & 100 & 0  & 0  & 34 \\ \hhline{-~-----}
Global  &                                            & 180\degree & 100 & 0  & 0  & 24 \\ \hline
Local      & \multirow{2}{*}{$1\textsubscript{Fr} 2\textsubscript{Sen}$}       & 180\degree & 50  & 50 & 0  & 21 \\ \hhline{-~-----}
Global  &                                            & 180\degree & 70  & 30 & 0  & 12 \\ \hline
\multicolumn{3}{|c|}{Local Average} & 83.75 & 11.25 & 5 & 35 \\ \hline
\multicolumn{3}{|c|}{Global Average} & 93.75 & 6.25 & 0 & 28 \\ \hline
\multicolumn{3}{|c|}{All Average} & 88.75 & 8.75 & 2.5 & 31 \\ \hline
\end{tabular}
\caption{Results from all classifiers (SVM), grasps and strategies when testing in real time the slip detection capabilities of the T-MO hand. TP: True positive, FP: False positive (reacted before slip), FN: False negative (item dropped) all (\%). D is average slipping distance in mm.}
\label{tab:hand_online}
\end{table}
%%%%%%%%%%%%%%%%%%%%%%%%%%%%%%%%%%%%%%%%%%%%%%%%%%%%%%%%

\subsubsection{Online Performance} \label{sec:hand_online_pringles}
The first online test was performed on a single object (the crisps can) to determine how well the offline results work in real time and to develop a strategy to respond to the onset of slip. The test conditions are the same as the single finger case: the object is grasped and then slowly released until slip occurs. \textcolor{black}{When slip is detected, the fingers are closed more tightly than at the start to counteract the momentum of the falling object (2\% of the finger motion range; chosen to add more force but not grasp the object too tightly)}. Again, we find the object height with an ArUco marker to record falling distance and identify false positives.

Our main aim is to determine which of the two classifiers developed under offline conditions works best online: the \textbf{global} (trained on data from all sensors) or the \textbf{local} (three classifiers each trained on a separate sensor). To do this we tested two different grasps: the first is the same as the training data; the second has the hand rotated through $180\degree$ (Fig.~\ref{fig:all_grasps}; hereafter referred to as $\mathbf{0\degree}$ and $\mathbf{180\degree}$ grasps). The hand rotation redistributes the forces on the second and third fingers, which will indicate which classifier is more robust to variation.

%%%%%%%%%%%%%%%%%%%%%%%%
\begin{figure}[t]
  \centering
    \text{Slip Detection Scores for Grasps \& Classifiers}
    \includegraphics[width=0.485\textwidth]{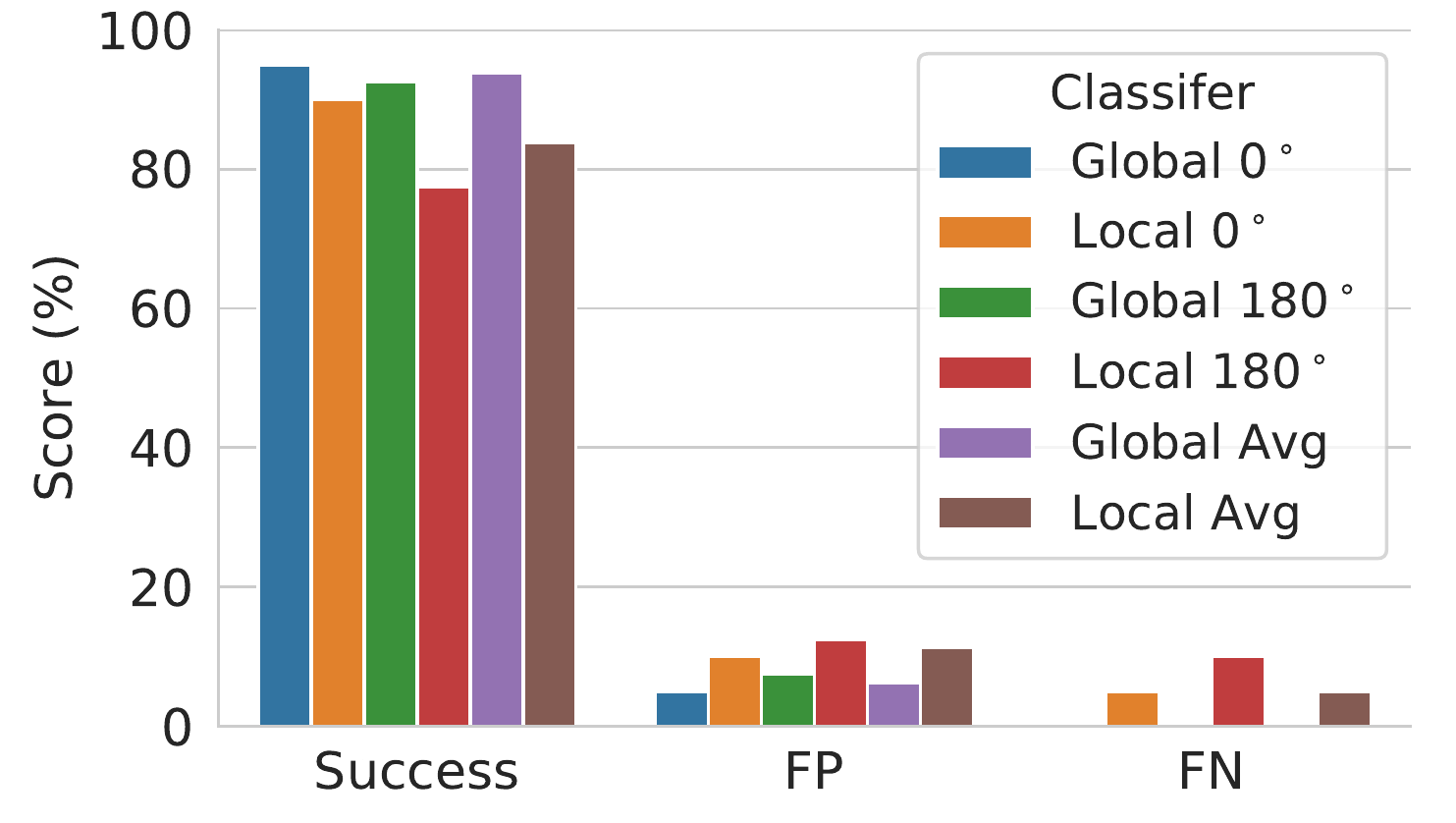}
    \caption{Comparison of successes, false positives (FP) and false negatives (FN) when reacting to the onset of slip with two classifiers in two orientations. Both classifiers score highly but the local classifier's inability to generalise to the rotated orientation makes the global classifier the better choice going forward.}
    \label{fig:hand_online_scores}
\end{figure}  
%%%%%%%%%%%%%%%%%%%%%%%%%

An additional aim is to determine how changing the slip response strategy affects performance. As described with the offline experiments (Section \ref{sec:hand_exp_des}), there are two parameters: the number of consecutive frames to be classified as slip and the number of sensors that simultaneously detect slip. It was quickly determined that using just a single sensor to trigger slip would lead to poorer performance compared with waiting for $2$ or $3$ to slip; therefore, this case was omitted from further testing. The remaining four strategies used are:
\begin{enumerate}
    \item Two consecutive frames, three sensors, $2\textsubscript{Fr} 3\textsubscript{Sen}$.
    \item Two consecutive frames, two sensors, $2\textsubscript{Fr} 2\textsubscript{Sen}$.
    \item One frame, three sensors, $1\textsubscript{Fr} 3\textsubscript{Sen}$.
    \item One frame, two sensors, $1\textsubscript{Fr} 2\textsubscript{Sen}$.
\end{enumerate}
These response strategies will henceforth be referred to by the code in the final column of this list. This was tried for both grasps (${0\degree}$, ${180\degree}$), giving a total of $8$ scenarios per classifier.

\begin{figure}[t]
  \centering
  \text{Slip Distances for Grasps Classifiers \& Strategies}
    \includegraphics[width=0.485\textwidth]{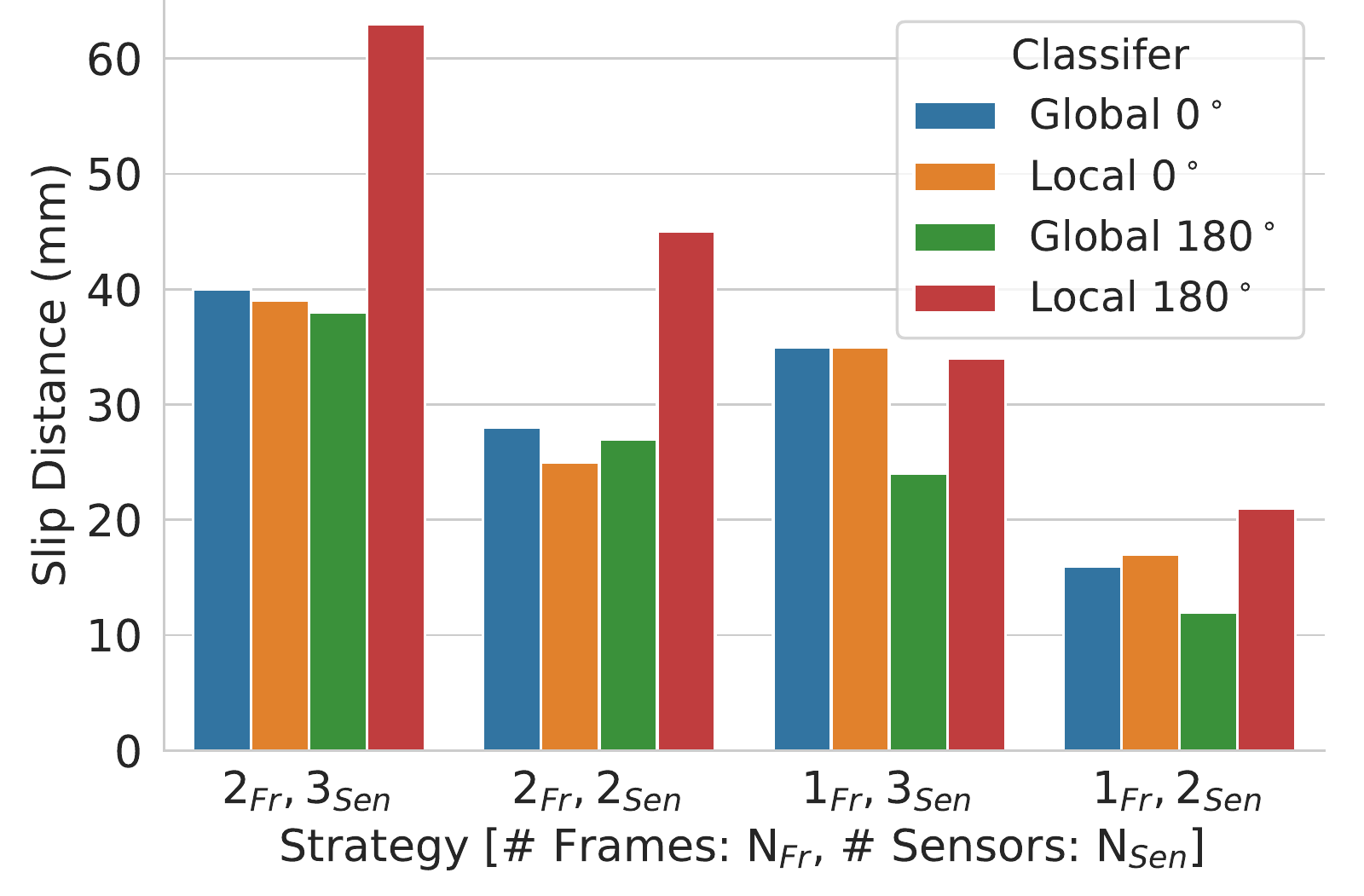}
    \caption{Slipping distance of the can for each strategy when regrasped after slip detected. The local classifier struggles in the rotated orientation as it fails to generalise to different grasps. The most sensitive strategy ($1\textsubscript{Fr} 2\textsubscript{Sen}$) has the smallest slipping distance as it requires the least evidence before it acts to minimise slip.}
    \label{fig:hand_online_dists}
\end{figure}

Ten trials were performed for each scenario, with most scoring highly: three quarters scored $80\%$ or above, with many at $100\%$ (results summarised in Figs. \ref{fig:hand_online_scores} \& \ref{fig:hand_online_dists} and Table. \ref{tab:hand_online}). A trial is considered a success if the object is seen to fall briefly before the hand responds and reestablishes a stable grasp.

In general, false positives (reacting before slip visibly occurs) occur infrequently ($8.75\%$ across all trials, Fig. \ref{fig:hand_online_scores}) with all from the trials where only one frame was required to be considered a reliable slip signal and only two sensors simultaneously detected slip to trigger a response (strategy $1\textsubscript{Fr} 2\textsubscript{Sen}$). This is the most sensitive strategy so it is unsurprising that it leads to the occasional false positive. All the false negatives (object dropped) occurred with the local classifier when the hand is rotated. This is a different grasp to the training data collection making generalisation difficult, but the false positive rate is low ($2.5\%$ across all trials).
%%%%%%%%%%%%%%%%%%%%%%%%
%%%%%%%%%%%%%%%%%%%%%%%%%

When testing in the same grasp configuration as for training data collection (hand 0\degree) the classifiers score very similarly with scores of $95\%$ and $90\%$ for the global and local classifier respectively (Fig. \ref{fig:hand_online_scores}). When the hand is rotated (180\degree) the global classifier again scores well ($92.5\%$; just a $2.5\%$ drop). The local classifier performs more poorly ($77.5\%$), which explains the difference in classifier score when averaged across the $8$ trials: the global classifier ($93.75\%$) outscores the local ($83.75\%$) classifier (Table \ref{tab:hand_online}).

%%%%%%%%%%%%%%%%%%%%%%%%
\begin{figure}[t]
  \centering
    \text{Slip Detection Scores for Classifiers \& Strategies}
    \includegraphics[width=0.485\textwidth]{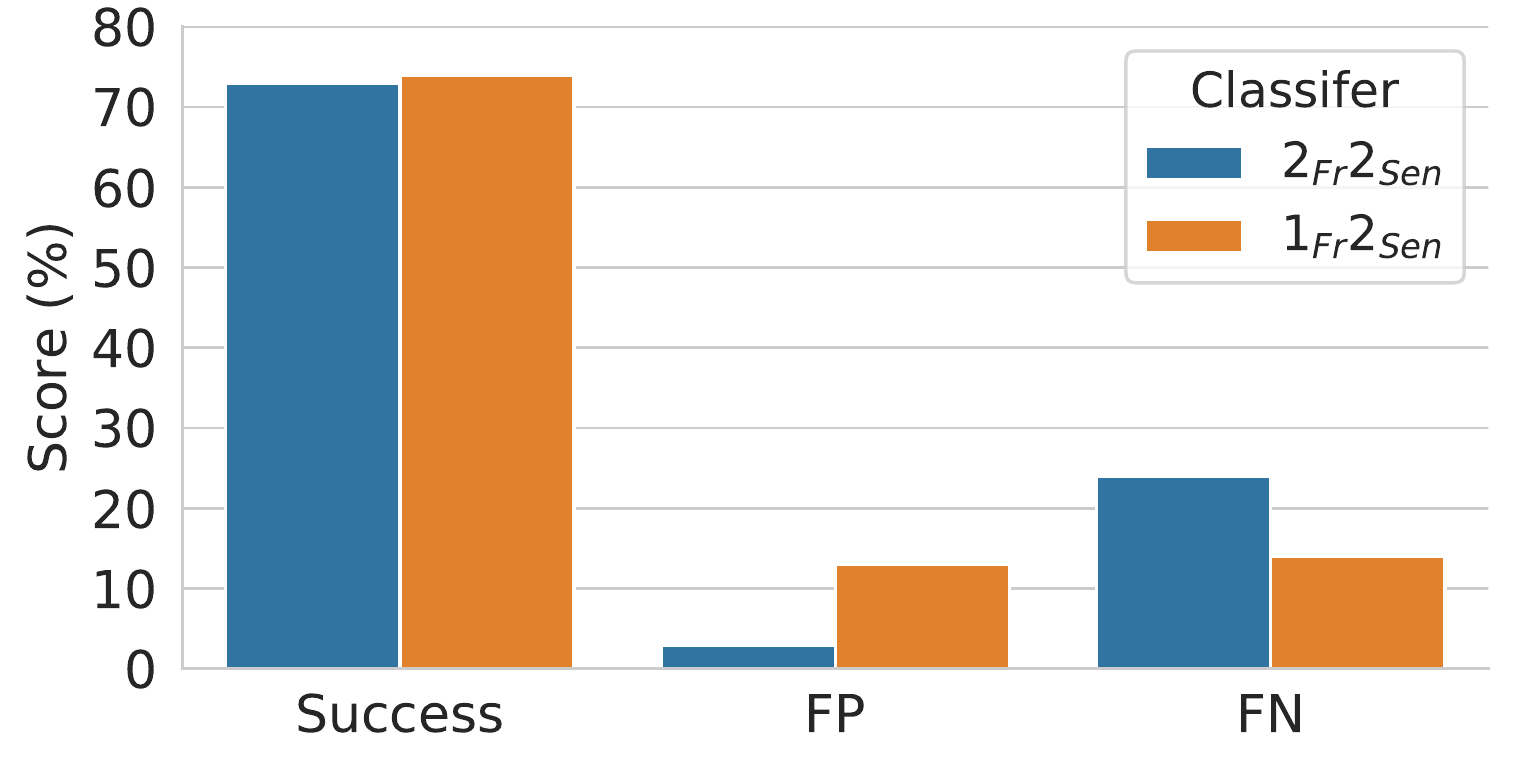}
    \caption{Average results for each strategy across all eleven objects (six of which were novel) when doing slip detection and grasp recovery.}
    \label{fig:hand_objects_bar}
\end{figure}  
%%%%%%%%%%%%%%%%%%%%%%%%%

%%%%%%%%%%%%%%%%%%%%%%%%
\begin{figure*}[t]
  \centering
    \begin{subfigure}[t]{0.32\textwidth}
        \centering
        (a) Establish Light Contact
        \includegraphics[width=\textwidth,trim={5cm 3cm 5cm 1cm}, clip=True]{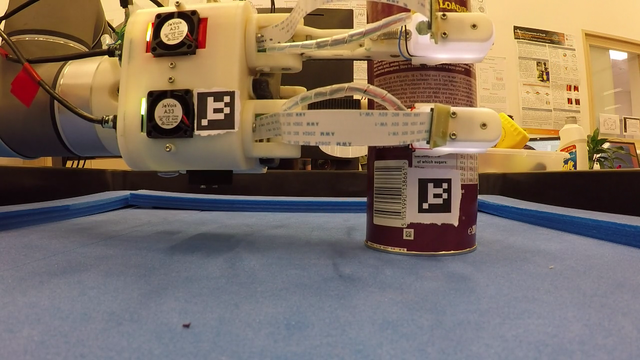}
    \end{subfigure}
        \centering
    \begin{subfigure}[t]{0.32\textwidth}
     \centering
    (b) Lift Arm; Object Slipping
        \includegraphics[width=\textwidth,trim={5cm 3cm 5cm 1cm}, clip=True]{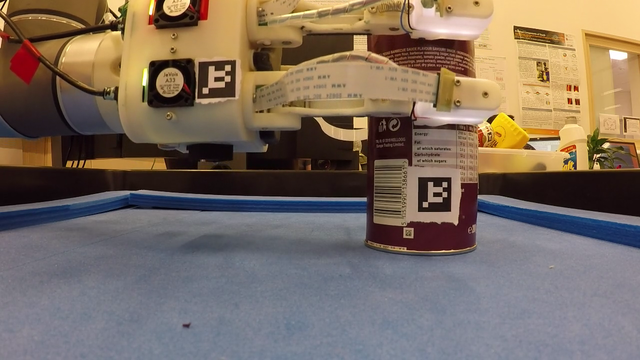}
    \end{subfigure}
    \begin{subfigure}[t]{0.32\textwidth}
     \centering
    (c) Grasp tightened; Object Grasped
        \includegraphics[width=\textwidth,trim={5cm 3cm 5cm 1cm}, clip=True]{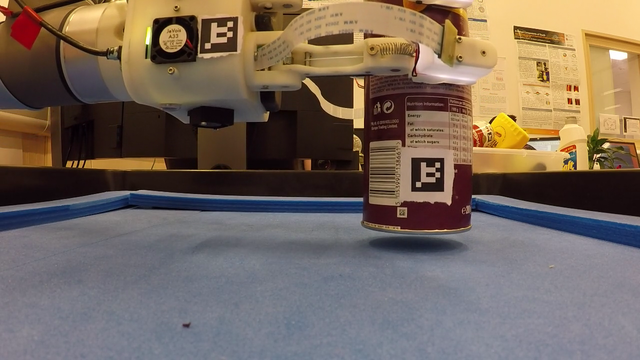}
    \end{subfigure}
    \caption{The stages of picking an object up first time using solely tactile data. (a) Initial contact is made with the object. (b) Arm is raised, slip is detected and hand is gripped tighter. (c) Sufficient force is applied and the object is securely lifted off the table. The difference in finger position on the can shows that the sensors did slip and the initial grasp was insufficient to lift the object. This verifies that the slip detection is sufficient to determine the minimum required force to lift an object of unknown weight.}
    \label{fig:pickup_flow}
\end{figure*}  
%%%%%%%%%%%%%%%%%%%%%%%%%

The slipping distances (Fig. \ref{fig:hand_online_dists}) show that the the global classifier performs well for both grasp orientations and all strategies, with similar slipping distances for both ($30$ mm normal, $25$ mm rotated). The local classifier in the normal orientation performs similarly; however, the performance drops significantly, for all but one of the strategies, when the hand is rotated $180\degree$ ($29$mm at $0\degree$, $41$mm at $180\degree$).

When each sensor only requires one frame classified as slip and only two sensors are required to trigger a response (strategy $1\textsubscript{Fr} 2\textsubscript{Sen}$), the slipping distance is much lower than the other strategies: $17$ mm across all trials for $1\textsubscript{Fr} 2\textsubscript{Sen}$ compared to the next best of $31$ mm for $2\textsubscript{Fr} 2\textsubscript{Sen}$ (Fig. \ref{fig:hand_online_dists}). This is, again, unsurprising as this strategy is the most sensitive and therefore reacts fastest to the onset of slip.

These results show that the global classifier outperforms the local classifier because it is able to generalise better to a different grasp; also, the more sensitive response strategies are more likely to result in false positives but will stop a slipping object quicker. For the remaining tests with the other four objects from the training dataset and the five novel objects, we will use the global classifier: a single SVM trained on data from all three sensors. We also have four different strategies that can be deployed depending on how sensitive we want the hand to be when detecting slip and initiating a response.

\subsubsection{Other and Novel Objects}
The final test was using the remaining four objects from the training dataset as well as six novel objects that are completely new to the classifier, two of which are deformable (Fig. \ref{fig:all_grasps}). We test strategies $2\textsubscript{Fr} 2\textsubscript{Sen}$ and $1\textsubscript{Fr} 2\textsubscript{Sen}$ from the previous test for each object with ten trials for each strategy. These are chosen as they gave the smallest slipping distance when regrasping an object (Fig. \ref{fig:hand_online_dists}). Results are presented in Table \ref{tab:hand_objects} and summarised in Fig. \ref{fig:hand_objects_bar}.

Both strategies score highly with $2\textsubscript{Fr},2\textsubscript{Sen}$ marginally outscoring $1\textsubscript{Fr},2\textsubscript{Sen}$ with successes of $74\%$ and $73\%$ respectively (Fig. \ref{fig:hand_objects_bar}). When $2\textsubscript{Fr},2\textsubscript{Sen}$ fails it is overwhelmingly due to it dropping the object (false negative) rather than reacting before slip occurs (false positive), achieving $3$\% FP, $24$\% FN across all trials. False positive and false negatives are much closer for $1\textsubscript{Fr},2\textsubscript{Sen}$, which has $13$\% FP, $14$\% FN across all trials. As $1\textsubscript{Fr},2\textsubscript{Sen}$ is the more sensitive strategy, these results are to be expected. $2\textsubscript{Fr},2\textsubscript{Sen}$ is less sensitive so requires more evidence before it will initiate a response; therefore, more objects being dropped is unsurprising.

A variety of grasps were used in this experiment (Fig. \ref{fig:all_grasps}), including two- and three-finger pinches and having the hand in different orientations. As the classifier was only trained with the hand in one orientation, this shows that the classifier is able to detect slip independent of hand orientation.
\begin{table}[t]
\centering
\begin{tabular}{|l|l|l|l|l|l|l|l|}
\hline
Object & Strategy   & TP  & FP & FN \\ \hline
\multirow{2}{*}{Ball}       & $2\textsubscript{Fr} 2\textsubscript{Sen}$     & 100 & 0  & 0  \\ \hhline{~----}
                            & $1\textsubscript{Fr} 2\textsubscript{Sen}$      & 90 & 10  & 0  \\ \hline
\multirow{2}{*}{Banana}     & $2\textsubscript{Fr} 2\textsubscript{Sen}$   & 90 & 0  & 10  \\ \hhline{~----}
                            & $1\textsubscript{Fr} 2\textsubscript{Sen}$    & 60 & 20  & 20  \\ \hline
\multirow{2}{*}{Brick}      & $2\textsubscript{Fr} 2\textsubscript{Sen}$    & 30 & 0  & 70  \\ \hhline{~----}
                            & $1\textsubscript{Fr} 2\textsubscript{Sen}$     & 50 & 0  & 50  \\ \hline
\multirow{2}{*}{Coffee}     & $2\textsubscript{Fr} 2\textsubscript{Sen}$   & 100  & 0 & 0  \\ \hhline{~----}
                            & $1\textsubscript{Fr} 2\textsubscript{Sen}$    & 50  & 50 & 0  \\ \hline
\multirow{2}{*}{Jello}      & $2\textsubscript{Fr} 2\textsubscript{Sen}$    & 20  & 0  & 80 \\ \hhline{~----}
                            & $1\textsubscript{Fr} 2\textsubscript{Sen}$     & 60 & 10  & 30  \\ \hline
\multirow{2}{*}{Mustard}    & $2\textsubscript{Fr} 2\textsubscript{Sen}$  & 90  & 0  & 10 \\ \hhline{~----}
                            & $1\textsubscript{Fr} 2\textsubscript{Sen}$    & 90 & 10  & 0  \\ \hline
\multirow{2}{*}{Bleach}     & $2\textsubscript{Fr} 2\textsubscript{Sen}$  & 100 & 0  & 0  \\ \hhline{~----}
                            & $1\textsubscript{Fr} 2\textsubscript{Sen}$    & 70 & 30  & 0  \\ \hline
\multirow{2}{*}{Soup}       & $2\textsubscript{Fr} 2\textsubscript{Sen}$     & 100  & 0 & 0  \\ \hhline{~----}
                            & $1\textsubscript{Fr} 2\textsubscript{Sen}$      & 100  & 0 & 0  \\ \hline
\multirow{2}{*}{Windex}     & $2\textsubscript{Fr} 2\textsubscript{Sen}$        & 60 & 30 & 10  \\ \hhline{~----}
                            & $1\textsubscript{Fr} 2\textsubscript{Sen}$        & 100 & 0 & 0   \\ \hline
\multirow{2}{*}{Pringles}   & $2\textsubscript{Fr} 2\textsubscript{Sen}$      & 100 & 0 & 0   \\ \hhline{~----}
                            & $1\textsubscript{Fr} 2\textsubscript{Sen}$      & 80 & 20 & 0   \\ \hline
\multirow{2}{*}{Cat (66g)}     & $2\textsubscript{Fr} 2\textsubscript{Sen}$        & 30 & 0 & 70  \\ \hhline{~----}
                            & $1\textsubscript{Fr} 2\textsubscript{Sen}$        & 60 & 0 & 40   \\ \hline
\multirow{2}{*}{Cat (166g)}   & $2\textsubscript{Fr} 2\textsubscript{Sen}$      & 60 & 0 & 40   \\ \hhline{~----}
                            & $1\textsubscript{Fr} 2\textsubscript{Sen}$      & 80 & 0 & 20   \\ \hline
\end{tabular}
\caption{Results from all objects. TP: True positive, FP: False positive (reacted before slip), FN: False negative (item dropped) all (\%).}
\label{tab:hand_objects}
\end{table}
%%%%%%%%%%%%%%%%%%%%%%%%

Some objects perform worse than others: notably, the sponge brick and the cat (Table \ref{tab:hand_objects}). \textcolor{black}{These are both deformable objects, which indicataes that a more complex response strategy may be needed to deal with such objects. However, when a 100g mass was attached to the cat the score raised from $45\%$ to $70\%$ across all trials}. The coffee, windex, bleach and banana score highly with one method ($\geq 80\%$) but struggle with the other. The remaining objects score highly when using both strategies. The objects in the training set score higher ($84\%$) than the novel objects ($70\%$); however, this is skewed by the Jello, which is thin so contact area is small, and the brick, which is very light and compliant.

\section{Application Scenarios}
\subsection{Experiment Description}
The motivation for developing an effective slip detection method using the Tactile Model O (T-MO) was to use the classifier in real scenarios where slip detection can be used to improve grasping performance or the effectiveness of the hand. To test this for the T-MO, we deploy the slip detection capabilities in two scenarios.

\subsubsection{Grasp Destabilisation}
%%%%%%%%%%%%%%%%%%%%%%%%
\begin{figure}[b]
  \centering
    \begin{subfigure}[t]{0.24\textwidth}
        \includegraphics[width=\textwidth,trim={0cm 0cm 0cm 0cm},clip=true]{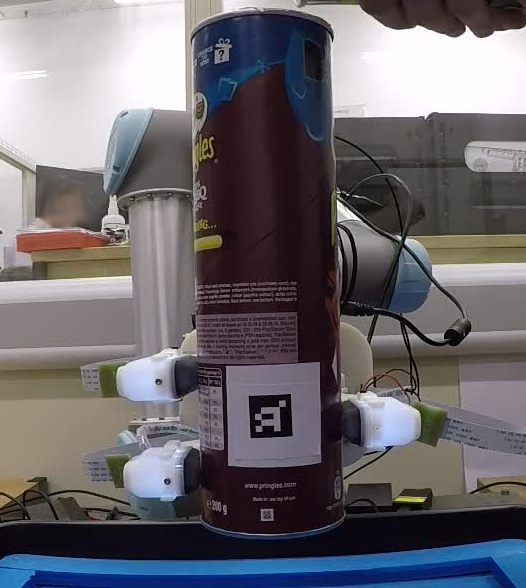}
    \end{subfigure}
        \centering
    \begin{subfigure}[t]{0.24\textwidth}
        \includegraphics[width=\textwidth,trim={0cm 0cm 0cm 0cm},clip=true]{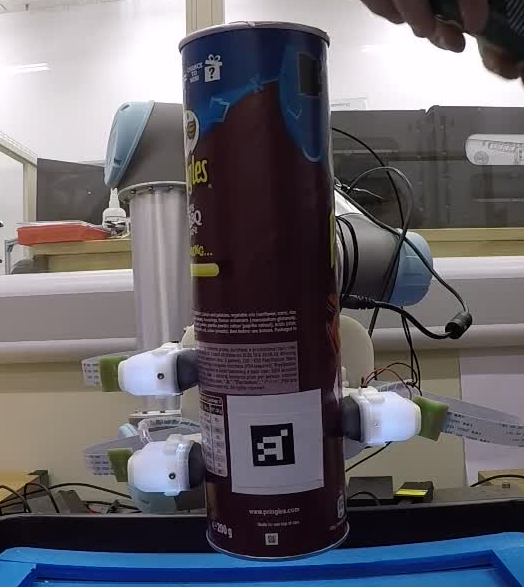}
    \end{subfigure} %\\[-1ex]
    \caption{Grasping object before (L) and after (R) adding rice to destabilise. Slip is detected and the object is caught after slipping a short distance.}
    \label{fig:destabilise}
\end{figure}  
%%%%%%%%%%%%%%%%%%%%%%%%%
Firstly, we use it to adjust the grasp strength when weight is being added to an object that is already securely held. This is a situation that humans encounter often and handle with ease, such as when liquid is being poured into a glass. By detecting when slip is occurring the T-MO can increase the grasp strength to compensate for the added weight.

For obvious reasons, we avoided using water to add weight, and instead poured rice into the empty crisps can that had been used in previous tests. This experiment gives a more gradual increase in weight rather than dropping larger masses in one by one, which could instead cause a slip but not make the object fall (and would thus incorrectly appear as though the slip detection is stopping a fall). 

\subsubsection{First Time Grasping}
The second scenario is to use slip to find a minimal grasp strength for lifting an object. In reality, most solid objects are unlikely to be damaged by existing robot hands on full power, which are far weaker than the human hand; hence, one strategy would be to establish a candidate grasp, squeeze as tightly as possible, and hope it is stable. However, this strategy would be disastrous when handling fragile or deformable objects (soft food, thin plastic cups, etc), where finding the sweet spot between securely grasping an object and not damaging it is essential. 

Therefore, we close the hand until it is in contact with the object and then raise the UR5 arm. If slip is detected, the hand will increase its grip strength, which the cycle repeated until the object is lifted, first time, with a minimal force grasp. This allows the T-MO to lift objects of unknown weight first time using only tactile information (Fig. \ref{fig:pickup_flow}). 

To establish a grasp with no prior knowledge of the object and no vision system requires detection of when initial contact occurs. We use a crude measure of sensor deformation from the mean absolute difference between individual pin positions at time $0$ (no sensor contact) and time $t$: 
\begin{equation}\label{eq:deform}
    d(t) = \frac{1}{30}\sum_{p=1}^{30} \sqrt{(x_{p}(t)-x_{p}(0))^{2} + (y_{p}(t)-y_{p}(0))^{2}}
\end{equation}
where $x_{p}$ and $y_{p}$ are the Cartesian pixel coordinates of a pin.

Hence, the strength of an initial grasp can be set by slowly closing the hand until the deformation $d(t)$ reaches a desired value. This value should be set to deform the sensors sufficiently for slip to be reliably detectable but not so deformed that the object will be picked up on the first try without increasing the grip strength.

The other variables to consider are the slip detection strategy and the length of pause after triggering a response to slip. As the fingers are initially not contacting the object, we should use the most sensitive slip detection strategy; i.e. a single frame from a single sensor to trigger a response $1\textsubscript{Fr} 1\textsubscript{Sen}$. \textcolor{black}{When slip is detected, the hand is closed a small amount corresponding to $1\%$ of its entire range, which is small enough to cause increments that will not result in much over-grasping while minimising the number of slips required to lift an object.} The time delay after a response was $0.1$s, which gives enough time for the hand to move and new data to be taken.

\textcolor{black}{Determining the amount of excessive force applied when lifting is important to ensure fragile/deformable objects are not crushed. As the T-MO has no force sensing, we estimate this by measuring the grasp position at which both objects could be lifted and compared it to the position when using slip detection. We consider the grasp strength as proportional to the increase in finger position between initial contact with the object and the grasp position, and thus define overgrasp as}

\textcolor{black}{\begin{equation} \label{eqn:overgrasp}
    o = \frac{(g_{\rm slip}-g_{\rm cont}) - (g_{\rm min}-g_{\rm cont})}{g_{\rm min}-g_{\rm cont}} = \frac{g_{\rm slip}-g_{\rm min}}{g_{\rm min}-g_{\rm cont}},
\end{equation}}

\noindent where $g_{\rm cont}$, $g_{\rm min}$ and $g_{\rm slip}$ are the grasp positions for contact with the object, minimum position to lift the object and the position obtained using slip detection to select the grasp. \textcolor{black}{$g_{\rm min}$ was determined for each object mass by systematically increasing the grasping strength until the object could be lifted.}

To verify that the object was successfully lifted and that the initial grasp was insufficient to lift the object, we attached an ArUco marker to both the object and the T-MO to track both 00their heights throughout each experiment.

\subsection{Results}
\subsubsection{Grasp Destabilisation}
The first experiment, where rice was added to an empty can, was successfully repeated five times. Here we used strategy $2\textsubscript{Fr} 2\textsubscript{Sen}$, which cuts a balance between avoiding false positives and having a small slipping distance. A clear slippage was noticed prior to the hand responding and the can was never dropped. The only negative was that in two of the trials a small slippage was visible when no response was triggered, then a larger slippage resulted in the hand gripping tighter. However, as the success criteria is the object not being dropped, this did not affect task success. 

Figure \ref{fig:destabilise} shows a grasped object before and after slip is induced and detected. A plot of the object height shows that the object moved slightly downwards as rice is added before a rapid drop in height is notable. This initial small drop is most likely due to the sensors sagging as they cope with the extra weight, rather than a false positive when slip is taking place.
%%%%%%%%%%%%%%%%%%%%%%%%
\begin{figure}[t]
  \centering
    \includegraphics[width=0.485\textwidth,]{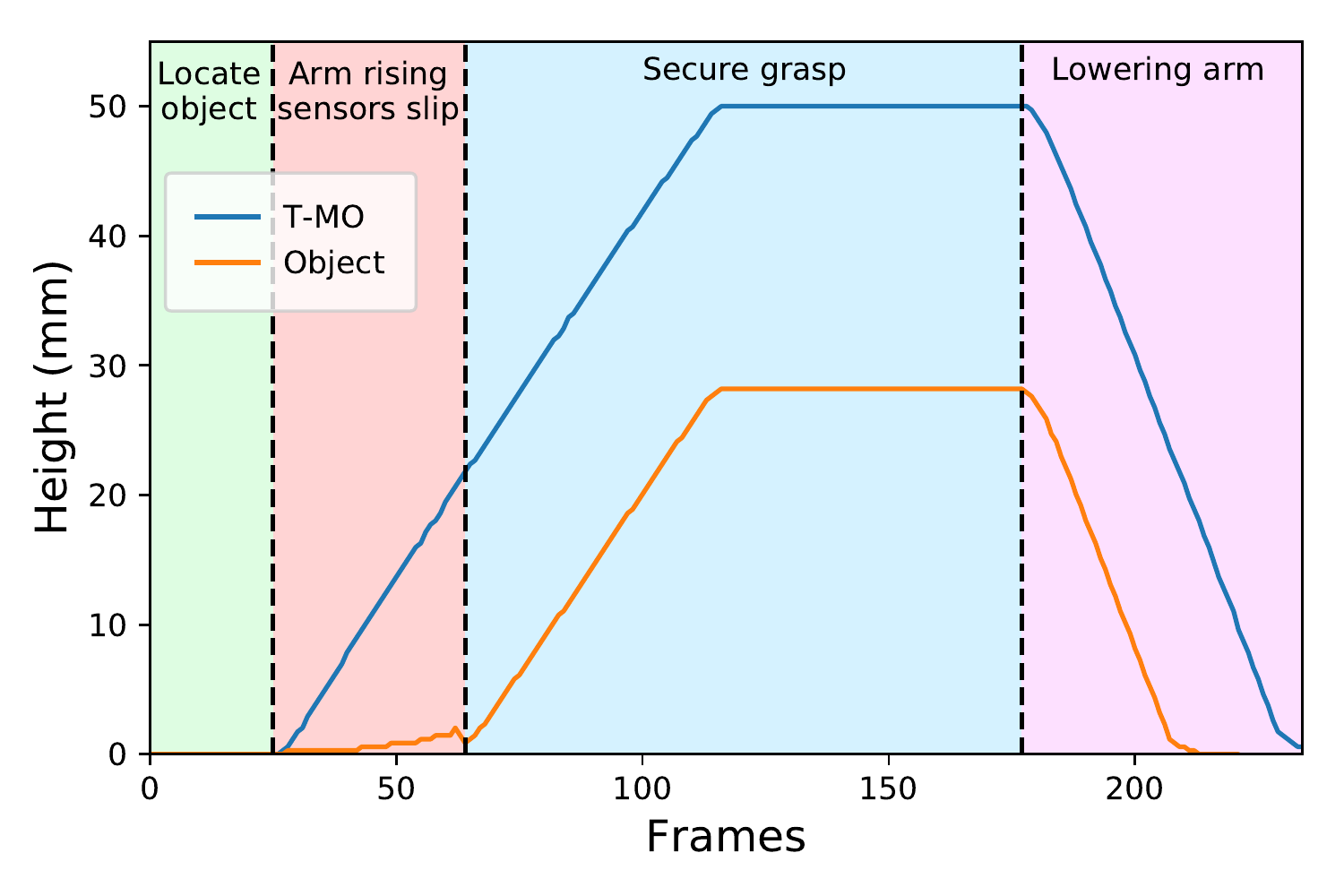}
    \caption{Height of the T-MO and the object above the ground when attempting to lift the object first time. The four stages indicated by background colour: (1) Hand makes initial contact (2) Arm lifts with insufficient contact force to lift object, slip is repeatedly detected and hand tightens (3) Object held (4) Object lowered. Heights recorded using ArUco markers (Fig. \ref{fig:pickup_flow}).}
    \label{fig:pickup_aruco}
\end{figure}  
%%%%%%%%%%%%%%%%%%%%%%%%%

\subsubsection{First-time Grasping}
In the second scenario, we seek to establish a stable grasp on the first attempt. Seven different crisps can masses from 50g (empty) to 500g were used with ten trials per mass. \textcolor{black}{We also tested a deformable object (the cat) both on its own and with a 100g mass attached. We set the deformation coefficient (Eq. \ref{eq:deform}) at 0.5, which was the minimum sensor deformation that reliably triggered slip}, and raised the hand at $17$mms$^{-1}$. To be successful, the hand must rise noticeably before the object (Fig. \ref{fig:pickup_aruco}). 

\begin{figure}[t]
  \centering
    \includegraphics[width=0.485\textwidth]{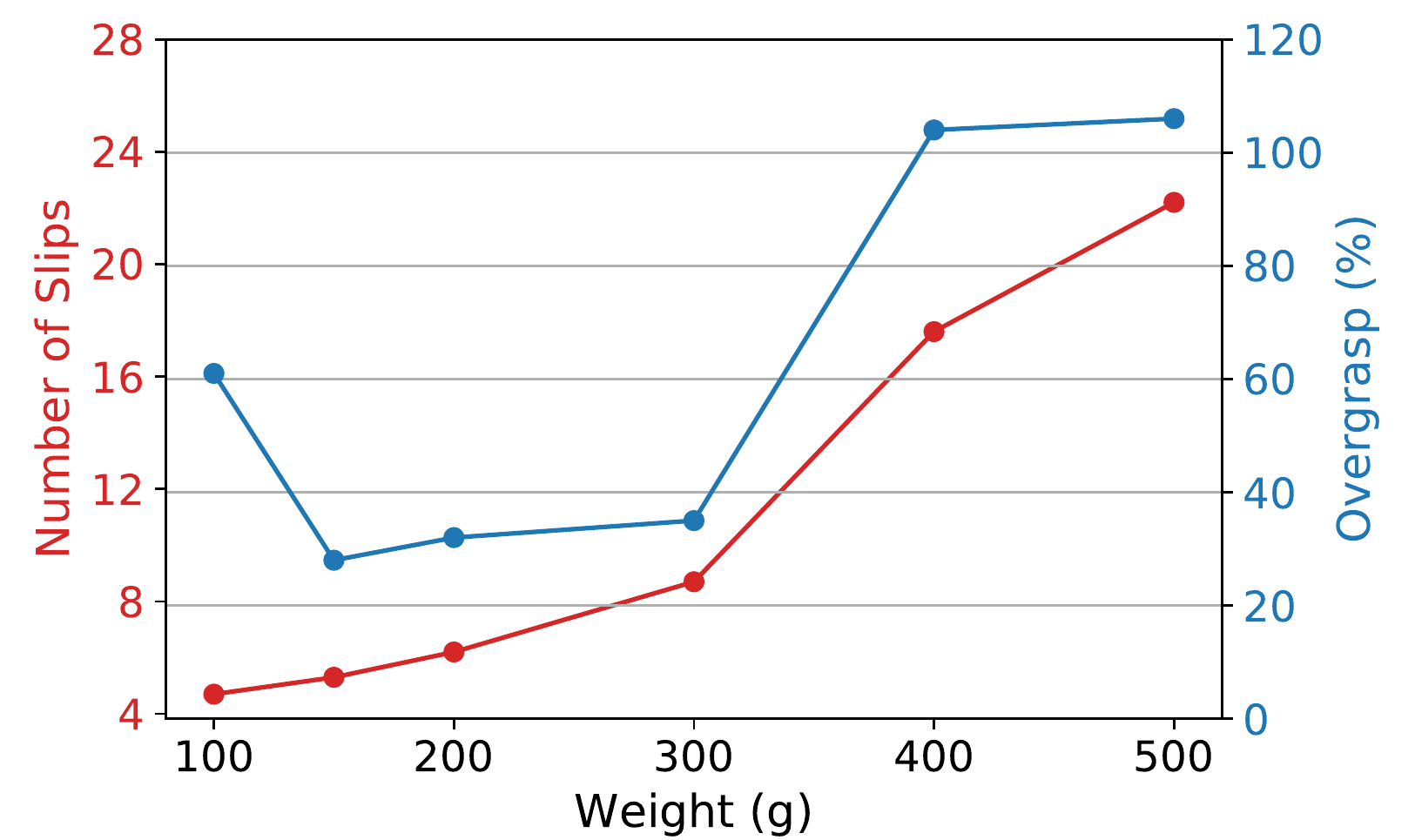}
    \caption{Number of slips detected (red) and overgrasp percentage (blue) when lifting an object at different masses. The overgrasp is an approximate measure of how excessively tight the grasp is when lifting. For low mass overgrasp is low showing that the object is picked without grasping too tight.}
    \label{fig:pickup_line}
\end{figure}

\textcolor{black}{The 50g crisps can was on the lower mass limit of this experiment, and we were not certain that the can slipped before being lifted. Each of the remaining weights and the deformable cat (at both weights) were successfully lifted first time over all ten trials. At masses between 100-300g the grasping position on the solid object was on average $39\%$ too tight (Fig. \ref{fig:pickup_line}, red line; using Eq. \ref{eqn:overgrasp}). At larger masses, the over-grasp becomes greater (105\%); the deformable cat was over-grasped by $34\%$ and $74\%$ when unweighted (66g) and weighted (166g) respectively.}

Intuitively, the heavier the object the more the hand would have to close to pick it up, and therefore more slips should be detected. This is reflected in the results (Fig. \ref{fig:pickup_line}, red line) where the number of independent slips detected (i.e. the number of times the hand gripped tighter) correlates with the mass of the object; therefore, the heavier the object, the stronger the hand is gripping. These results demonstrate that the T-MO is highly capable at picking up objects of unknown mass first time using only tactile data. 
%%%%%%%%%%%%%%%%%%%%%%%%
%%%%%%%%%%%%%%%%%%%%%%%%%

\section{Limitations and Future Work}
\subsection{Objects for Testing}
As investigated in James et al. (2020), the finger thickness on the Tactile Model O (T-MO) means that the hand struggles to pick up very small and flat object from a surface \cite{Church2019}. This restricts the number of objects on which we can perform tests. Given that the purpose of this paper is to test the slip detection capabilities of the T-MO, we deliberately selected objects that were previously determined as easily graspable, as there would be little point trying to test how well the T-MO could react to slip on an object that it could barely grasp. \textcolor{black}{However, given these objects, the method can only be considered as applicable to objects graspable with an antipodal pinch.}

We have demonstrated that the T-MO is highly capable of slip detection and grasp stabilisation, and have used these skills in tasks involving slip detection that encountered in real-world scenarios. This, coupled with the T-MO's strengths as a low-cost manipulator, provide a motivation to refine the design to improve grasping performance and subsequently widen the range of objects on which slip could be detected in the future.  

\subsection{Classification Method}
One of the advantages of the JeVois onboard processing units encased within the T-MO is their ability to reduce computational load on the control PC. By processing their data into pin positions, the PC does not need to process three high bandwidth video streams and can communicate with the JeVois over low-bandwidth serial ports.

Prior work with the T-MO (when using raw images) operated at a low frame rate (20fps) to allow for high resolution images to be captured \cite{Church2019}, but this had significant latency when using a deep convolutional neural network. We did not use those methods here as the high frame rate (60fps) and low latency from using pin positions is beneficial when responding to events that happen on a very short time scale. We have demonstrated that pin positions provide an effective tactile signal for slip detection and gives the T-MO hand the ability to perform grasping tasks. Further work exploring the use of high frame rate image capture and using the JeVois' onboard processing to run small TensorFlow Lite models on images would be interesting but are beyond the scope of this study. 

\subsection{Incipient Slip and Slip Prediction}
A significant area of research within tactile sensing is the detection of incipient slip, where part of the contact surface is slipping and part is static; as opposed to gross slip, where the entire contact surface is slipping \cite{Chen2018}, which has been investigated here. \textcolor{black}{Incipient slip is a complicated phenomenon that is influenced by the surface properties of the tactile sensor and the object being held, including the coefficient of friction and surface geometry. Depending on conditions, it may only be present on a short timescale, which means the ability to detect incipient slip depends on the frame rate of the tactile sensor as well as the properties of the sensor's surface}. A preliminary examination of the data did not indicate the presence of incipient slip, and reliable detection of incipient slip with the T-MO would likely involve significant hardware modifications.

Here, using gross slip detection, we have demonstrated strong performance at grasping objects, preventing novel objects from being dropped, and presented several control strategies that can be used depending on the task. None of these relied on incipient slip. We anticipate that making the necessary changes required to detect incipient slip would enhance performance and we will investigate this in the future. 

\section{Discussion and Conclusion}
We have presented a method of using a three-fingered hand with integrated optical tactile sensors -- the T-MO -- to perform slip detection. Using various simple slip response strategies, we grasped objects at the first attempt, performed slip detection on novel objects and regrasped slipping objects. 

\textcolor{black}{To justify the appropriate slip detection method, we considered several approaches that maintained an interpretation in terms of the pin velocities being a direct measure of slip occurrence: first, a threshold on the magnitude of mean pin velocity; second, a support vector machine applied to either linear or nonlinear combinations of pin velocities; and third a logistic regression (LogReg) method applied to individual pin velocities. The  threshold  method  performed poorly due to the optimal threshold significantly overlapping with the slip class (Fig.~4), which appears to be because of changes of contact that do  not cause  slip yet still result in large pin velocities. A more sophisticated classifier is therefore necessary to capture the complexities of the slip/static boundary. While both the SVM and LogReg performed well across all tests  (Fig.~5), the  logistic regression dropped in performance when down-sampling data. Hence, the  SVM was the  superior method  for  detecting slip and was used in all further analyses.}

A related prior study by James et al. (2018) \cite{James2018} demonstrated that a larger TacTip could perform slip detection and react to catch an object sliding in a rail (same setup as Fig. \ref{fig:slip_rig}) after only 16mm of falling. Under the same conditions we obtained distances of 21mm and 29mm depending on the strategy used to respond to slip. These reduced scores are most likely explained by the cameras here running at 60fps compared to 100fps in the TacTip and the flexibility of the T-MO's fingers leading to a reduced force being applied to stop a falling object. Nevertheless, the T-MO's demonstrated ability to successfully stop slipping objects within a short distance and the ability to deploy slip detection in realistic scenarios represents a significant advance.

We tested the slip detection capabilities of the entire hand by regrasping eleven slipping objects, six of which were novel and one highly deformable, using various strategies. \textcolor{black}{Li et al. (2017) also use an SVM to detect slipping objects using a three fingered hand, but only use this to detect slip but do not regrasp slipping objects} \cite{li2017learning}. Here, a classifier trained on data from all three sensors was shown to be robust to changes in grasp and was able to regrasp a crisps can after a slipping distance of 12mm (70\% success, 30\% FP) and 24mm (100\% success) depending on the strategy used. False positives are better than false negatives as the object could be dropped; however, minimising these is still desired. \textcolor{black}{Using a force meter the peak force of the slip response was measured to be \textasciitilde $3$N, which was less than the forces to crush a cardboard tube (4N), a strawberry (3.5N) or an eggshell (20N) with the sensor; thus, the slip response is unlikely to damage delicate objects.}

Scaling up from one to three sensors represents a significant challenge as the complexity of the interaction between the object and the sensors increases greatly. This means that there is no guarantee that the features present in the single-finger case will be prominent or even present in the three-finger case. Additionally, the robot hand must be capable of collecting tactile data, classifying it and reacting quickly. Therefore, demonstrating that the T-MO is capable of preventing a slipping object from being dropped is the culmination of two distinct contributions.

The final task was to apply the developed classifier in two scenarios: preventing an object being dropped when weight is added and picking an object up without excessive force. The ability to cope with added weight is an important feature of a gripper; for example, the grasped item may be a box being filled with objects or a glass being filled with water. Heydarabad et al. (2017) performed a similar experiment but added weights in larger increments than the small rice grains used here \cite{Heydarabad2017High-performingControl}. They successfully detected slip and prevented the object from being dropped; however, dropping a mass to induce slip gives a much greater signal than gradually adding weight in the form of rice grains.

Grasping an object first time without excessive force is an important consideration as it minimises energy use, increases longevity and does not damage the object. We were able to pick up both solid and deformable objects at different weights first time. At masses between 100-300g, the grasping position on the solid object was on average $39\%$ too tight. The deformable object was overgrasped by $54\%$ across both weights. Humans have been shown to add a safety margin of between 10-40\% when preventing objects from slipping \cite{Johansson2008TactileHumans}, which our results do not greatly exceed (although we do acknowledge that our over-grasp measure is a basic approximation of excessive force). We anticipate that a bespoke classifier for lifting objects would perform better, but we have demonstrated that objects will likely be undamaged when lifted by the T-MO. A similar experiment was performed by Stachowsky et al. (2015), who successfully lifted a deformable cup ten times without damage \cite{stachowsky2016slip}, but in contrast only performed this with a single cup weight.

In conclusion, we have demonstrated that the Tactile Model O is a highly capable slip-detection platform when using a support vector machine as a classifier. The T-MO was able to detect slip in novel objects using an antipodal pinch grasp, and was robust to changes in grasp and slipping direction. We also developed different strategies for reacting to slip detection, of use in different scenarios. Our classifier was used in two realistic scenarios where the ability to detect slip allowed the T-MO to pick up an object of unknown weight with the approximate minimum necessary force and prevent it from being dropped when the weight was changed. These experiments also demonstrated that the ability to detect slip serves as a useful and robust metric for determining grasp stability. Overall, this shows the Tactile Model O (T-MO) is a suitable platform for a wide variety of autonomous grasping scenarios, by using its reliable slip detection capabilities to ensure a stable grasp in unstructured environments. 

\section*{Acknowledgements}
We thank John Lloyd, Kirsty Aquilina, Ben Ward-Cherrier, Luke Cramphorn, Nicholas Pestell, Andy Stinchcombe and Gareth Griffiths for their help throughout this work. 

\bibliographystyle{ieeetr}
\bibliography{jamesBibliography}
\begin{wrapfigure}{l}{0.13\textwidth}
\vspace{-2mm}
\includegraphics[width=0.98\linewidth]{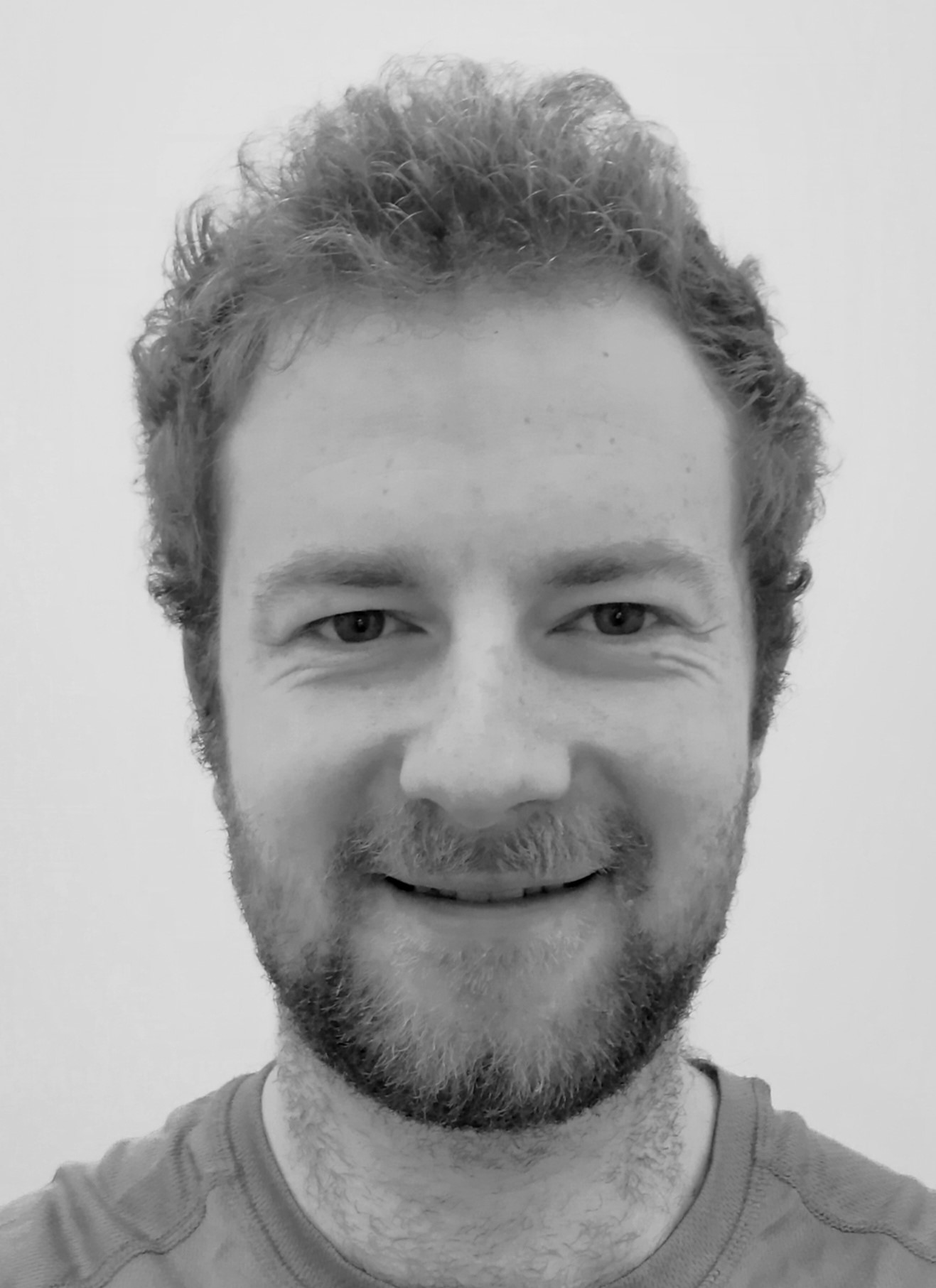}

\vspace{5.5mm}

\includegraphics[width=0.98\linewidth]{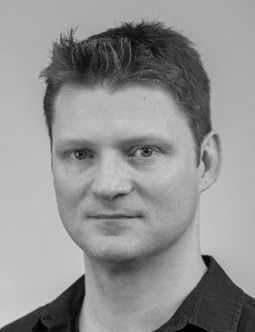}
\vspace{-10mm}
\end{wrapfigure}

\textbf{Jasper W. James} (Student Member, IEEE) received an M.Phys in Physics from the University of Oxford in 2012. He is currently completing a Ph.D. in Robotics at the University of Bristol focussing on slip detection with biomimetic optical tactile sensors and soft tactile robot hands. \par
\bigskip
\smallskip
\textbf{Nathan F. Lepora} (Member, IEEE) received the B.A. degree in Mathematics and the Ph.D. degree in Theoretical Physics from the University of Cambridge, UK. He is currently a Professor of Robotics and AI with the University of Bristol, UK. He leads the Tactile Robotics Group that researches human-like dexterity with soft tactile robots and AI such as deep learning.

He is a recipient of a Leverhulme Research Leadership Award on ‘A Biomimetic Forebrain for Robot Touch’. His research group won the ‘Contributions in Soft Robotics Research' category in the 2016 Soft Robotics Competition. He co-edited the book ‘Living Machines’ that won the 2019 BMA Medical Book Awards (basic and clinical sciences category). 
\end{document}